\documentclass[10pt,journal,cspaper,compsoc]{IEEEtran}
\ifCLASSOPTIONcompsoc
\else
\fi
\ifCLASSINFOpdf
\else
\fi
\ifCLASSOPTIONcompsoc
\usepackage[normalsize,sf]{subfigure}
\else
\usepackage[footnotesize]{subfigure}
\fi

\ifCLASSOPTIONcompsoc
  \usepackage[caption=false,font=normalsize,labelfont=sf,textfont=sf]{subfig}
\else1
  \usepackage[caption=false,font=footnotesize]{subfig}
\fi
\usepackage{times}
\usepackage{epsfig}
\usepackage{graphicx}
\usepackage{amsmath}
\usepackage{amssymb}
\usepackage{multirow}
\usepackage[algoruled,vlined,linesnumbered]{algorithm2e}
\usepackage{rotating}
\usepackage{bbm}
\usepackage{bm}
\usepackage{color}
\usepackage{url}
\usepackage{threeparttable}

\usepackage{algorithmic}

\usepackage[nospace,nocompress]{cite}

\usepackage{cancel}

\usepackage{algorithmic}

\DeclareMathAlphabet{\mathpzc}{OT1}{pzc}{m}{it} 

\newcommand{\etal}{et al.\!}

\newcommand{\ie}{i.e.\!}

\begin{document}

%
\title{Person Re-Identification by Discriminative Selection in Video Ranking}
%
%
%
%

\author{
Taiqing~Wang,
~Shaogang~Gong,
Xiatian~Zhu, 
and Shengjin Wang
\IEEEcompsocitemizethanks{
\IEEEcompsocthanksitem 
Taiqing Wang and Shengjin Wang are with State Key Laboratory of Intelligent Technology and Systems, Tsinghua National Laboratory for Information Science and Technology, Department of Electronic Engineering, Tsinghua University.
\protect\\
E-mail: wangtq09@mails.tsinghua.edu.cn, wgsgj@tsinghua.edu.cn
\IEEEcompsocthanksitem 
Shaogang Gong and Xiatian Zhu are with School of Electronic Engineering and Computer Science, Queen Mary University of London. 
\protect\\
E-mail: s.gong@qmul.ac.uk, xiatian.zhu@qmul.ac.uk 
\vspace{-0.05cm}
}
\thanks{This work was partially supported by National Science and Technology Support Program under Grant No. 2013BAK02B04 and Initiative Scientific Research Program of Tsinghua University under Grant No. 20141081253.}}

%
%

\markboth{
	}%
{T. Wang \MakeLowercase{\textit{et al.}}: Person Re-Identification by Video Ranking}
%


\IEEEcompsoctitleabstractindextext{%
\begin{abstract}
Current person re-identification (ReID) methods typically rely on single-frame imagery features, whilst ignoring space-time information from image sequences often available in the practical surveillance scenarios. Single-frame (single-shot) based visual appearance matching is inherently limited for person ReID in public spaces due to the challenging visual ambiguity and uncertainty arising from non-overlapping camera views where viewing condition changes can cause significant people appearance variations. In this work, we present a novel model to automatically select the most discriminative video fragments from noisy/incomplete image sequences of people from which reliable space-time and appearance features can be computed, whilst simultaneously learning a video ranking function for person ReID. Using the PRID$2011$, iLIDS-VID, and HDA+ image sequence datasets, we extensively conducted comparative evaluations to demonstrate the advantages of the proposed model over contemporary gait recognition, holistic image sequence matching and state-of-the-art single-/multi-shot ReID methods.

\end{abstract}

}

\maketitle

\IEEEdisplaynotcompsoctitleabstractindextext

%
\IEEEpeerreviewmaketitle

\section{Introduction}
\label{sec:intro}

\begin{figure}[!t]
\centering
\subfigure[\footnotesize Cross-view lighting variations]
{
\includegraphics[width=0.46\linewidth]{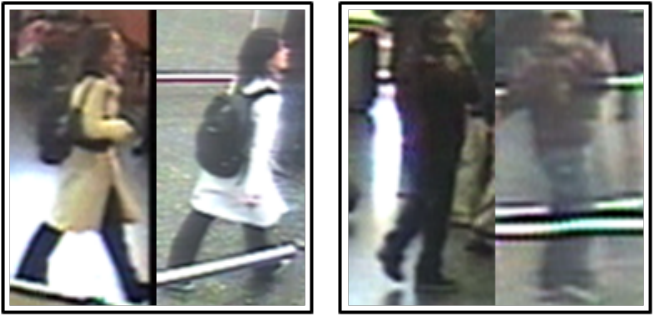}%
}
\hfil
\subfigure[\footnotesize Camera viewpoint changes]
{
\includegraphics[width=0.46\linewidth]{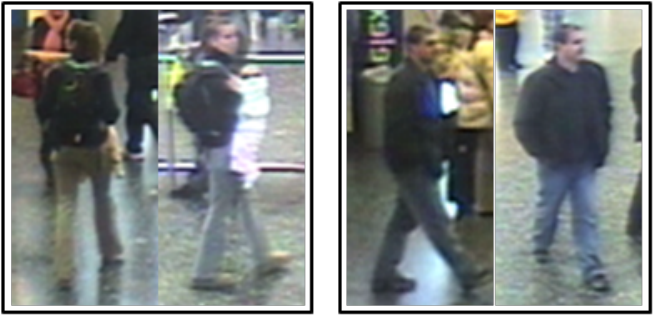}
}
\vskip -0.2cm
\subfigure[\footnotesize Clothing similarity]
{
\includegraphics[width=0.46\linewidth]{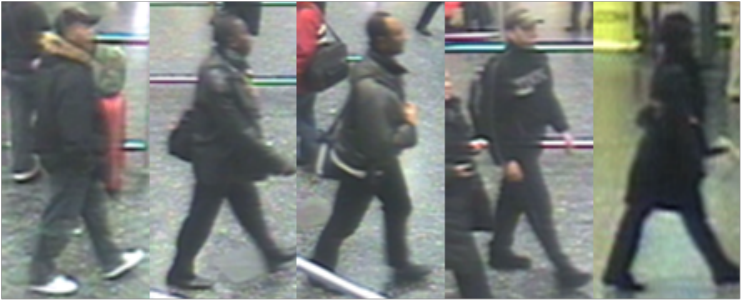}
}
\hfil
\subfigure[\footnotesize Background clutter/occlusions]
{
\includegraphics[width=0.46\linewidth]{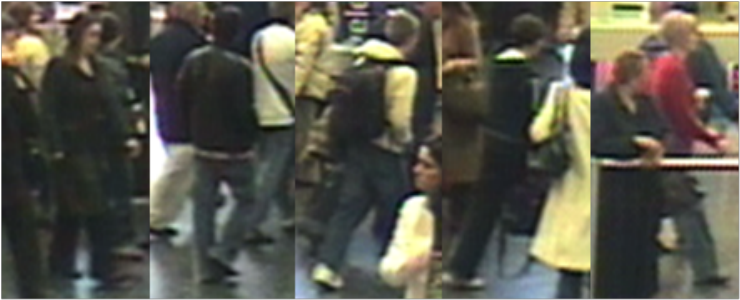}
}
\vskip -.3cm
\caption{\footnotesize
Person re-identification challenges in public space scenes~\cite{ilids_mctc08}.
(a-b) The two images in each bounding box refer to the same person 
observed in different cameras.
}
\label{fig:ambiguous_appearance}
\vspace{-.6cm}
\end{figure}

{\color{black}\IEEEPARstart{F}{or}
making sense of the vast quantity of video data generated by 
large scale surveillance camera networks in public spaces, 
automatically (re-)identifying individual persons
across non-overlapping camera views distributed at different physical locations is essential.
This task is known as \textit{person re-identification} (ReID). 
Automatic ReID enables the discovery and analysis of person-specific long-term
activities over widely expanded areas and
is fundamental to many important surveillance applications 
such as multi-camera people tracking and forensic search.
Specifically, for performing cross-view person ReID, 
one matches a probe (or query) person 
against a set of gallery people for generating a ranked list 
according to their matching similarity.
Typically, it is assumed that the correct
match is assigned to one of the top ranks,
ideally the top-$1$ rank~\cite{zheng2013re,hirzer2012relaxed,farenzena2010person,gheissari2006person, zheng2015scalable}.}
As the probe and gallery people are often captured from 
a pair of disjoint camera views at different times,
cross-view visual appearance variations
can be significant. 
Person ReID by visual matching is thus inherently
challenging~\cite{REIDChallenge:14}.
The state-of-the-art methods perform this task 
mostly by matching spatial appearance features
(e.g. colour and texture)
using a pair of single-shot person
images~\cite{hirzer2012relaxed,farenzena2010person,ProsserEtAlBMVC:10,zhao2013unsupervised}. 
However, single-shot appearance features of people are intrinsically
limited due to the inherent visual ambiguity caused by 
clothing similarity among people in public spaces
and appearance changes from cross-camera viewing condition variations 
(Fig.~\ref{fig:ambiguous_appearance}).
It is desirable to explore space-time information from image sequences
of people for ReID. 

Space-time information has been explored extensively for action
recognition~\cite{poppe2010survey,weinland2011survey}. Moreover,
discriminative space-time video patches have also been exploited for
action recognition~\cite{sapienza2012learning}. Nonetheless, action
recognition approaches are not directly applicable 
to person ReID because pedestrians in
public spaces exhibit similar walking activities without
distinctive and semantically categorisable action patterns unique to different
identities.

On the other hand, gait recognition techniques have been developed for person
recognition using image sequences by discriminating 
subtle distinctiveness in the style of
walking~\cite{nixon2010human,sarkar2005humanid}.
Different from action recognition, gait is a behavioural biometric that
measures the way people walk. An advantage of gait recognition is
no assumption being made on either subject cooperation (framing) or
person distinctive actions (posing). 
These characteristics are similar to person ReID
situations. However, existing 
gait recognition models are subject to stringent requirements on
person foreground segmentation and accurate alignment over time
throughout a gait image sequence (a walking cycle). 
It is also assumed that complete
gait/walking cycles were captured in the target image
sequences~\cite{Han2006GEI,martin2012gait}. Most gait
recognition methods do not cope 
well with cluttered background and/or random occlusions with
unknown covariate conditions \cite{BashirEtAl:PRL10}.
Person ReID 
is hence inherently challenging for gait
recognition techniques (Fig.~\ref{fig:ambiguous_appearance}).

In this study, we aim to construct a discriminative video
matching framework for person re-identification
by selecting more reliable space-time features from person videos,
beyond the often-adopted spatial appearance features. 
To that end, we assume the availability of
image sequences of people which may be highly noisy, 
i.e., with arbitrary
sequence duration and starting/ending frames, 
unknown camera 
viewpoint/lighting variations during each image sequence, 
incomplete frames due to uncontrolled occlusions,
\textcolor{black}{
	no guaranteed high frame rates,
	and possible clothing changes over time.
} 
We call these videos {\em
  unregulated} image sequences of people
(Fig.~\ref{fig:ambiguous_appearance} and
Fig.~\ref{fig:dataset_example}). 
More specifically, we propose a novel approach to
Discriminative Video fragment selection and Ranking (DVR) based on a
robust space-time and appearance feature representation given unregulated person image sequences. 

The main contributions of this study are:
(1) We derive a multi-fragment based appearance and space-time feature representation of
image sequences of people. This representation is based 
on a combination of HOG3D, colour and optic flow energy profile
of image sequence, designed to break down automatically unregulated
video clips of people into multiple fragments.
(2) We formulate a discriminative video ranking model for
cross-view person re-identification by simultaneously
selecting and matching more reliable appearance and space-time features 
from video fragments. The model is formulated using a multi-instance
ranking strategy for learning from pairs of image sequences over
non-overlapping camera views. 
The proposed method can relax significantly the strict assumptions made by
gait recognition techniques.
(3) We extensively provide comparative evaluations
of the proposed model against a wide range of contemporary methods 
(e.g. gait recognition, holistic sequence matching and 
state-of-the-art person ReID models)
on three challenging image sequence based datasets.

\section{Related Work}
\label{sec:related_work}

\noindent \textbf{Space-time features}
Space-time feature representations have been extensively explored in action/activity recognition~\cite{poppe2010survey,wang2009evaluation,gong2011visual}.
One common representation is constructed based on space-time interest
points
\cite{LaptevIJCV2005STIP,dollar2005behavior,willems2008efficient,bregonzio2009recognising}. 
They facilitate a compact description of image sequences based on sparse interest points,
but are somewhat sensitive to shadows and highlights in appearance~\cite{ke2010volumetric} and may lose discriminative information~\cite{gilbert2009fast}.
Therefore, they may not be suitable for person ReID scenarios where
lighting variations and viewpoints are unknown and uncontrolled.
Relatively, space-time volume/patch based representations
\cite{poppe2010survey} can be 
richer and more robust.
Mostly these representations are spatial-temporal extensions of 
corresponding image descriptors,
e.g. HoGHoF~\cite{laptev2008learning},
3D-SIFT~\cite{scovanner20073} and
HOG3D~\cite{klaser2008spatio}.
In this study, we adopt HOG3D~\cite{klaser2008spatio} 
as the space-time feature of video fragment 
because: 
(1) It can be computed efficiently; 
(2) It contains both spatial gradient and temporal dynamic information, 
and is therefore potentially more expressive~\cite{wang2009evaluation,klaser2008spatio};
(3) It is more robust against cluttered background and
occlusions~\cite{klaser2008spatio}. 
The choice of space-time feature is independent of our model.

\vspace{0.1cm}
\noindent \textbf{Gait recognition} 
Space-time information of sequences has been 
extensively exploited by gait
recognition~\cite{nixon2010human,sarkar2005humanid,Han2006GEI,martin2012gait}.
However, 
these methods often make stringent assumptions
on the image sequences, e.g. uncluttered background, consistent
silhouette extraction and alignment, accurate gait phase estimation and complete
gait cycles, most of which are unrealistic in ordinary person ReID scenarios. 
It is challenging to extract a suitable gait representation from typical
ReID data. 
In contrast, our approach 
relaxes significantly 
these assumptions by 
simultaneously selecting discriminative video
fragments from noisy sequences,
learning and matching them without temporal alignment.

\vspace{0.1cm}
\noindent\textbf{Temporal sequence matching} 
One approach to exploiting image sequences for
ReID is holistic sequence matching.
For instance, Dynamic Time Warping (DTW) is a popular sequence
matching method widely used for action recognition~\cite{lin2009recognizing},  
and recently also for person ReID~\cite{simonnet2012}.
However, given two unregulated sequences, 
it is difficult to align sequence pairs 
for accurate matching, 
especially when the image sequences are subject to significant noise caused by unknown
camera viewpoint changes, 
background clutter and drastic lighting changes. 
Our approach is designed to address this problem while avoiding any implicit
assumptions on sequence alignment and 
camera view similarity among image frames both within and between sequences.

\vspace{0.1cm}
\noindent \textbf{Multi-shot person re-identification} 
Multiple images from a sequence of the same person have been exploited for 
person re-identification. For example, interest points were accumulated across images for
capturing appearance variability~\cite{hamdoun2008person}, manifold
geometric structures in image sequences of people were utilised
to construct more compact 
spatial descriptors of people~\cite{cong2009video},
and the time index of image frames and identity consistency of a sequence were
used to constrain spatial feature similarity
estimation~\cite{karaman2012identity}. There were also attempts on
training a person appearance model from image sets~\cite{nakajima2003full} or
by selecting best pairs~\cite{li2013locally}.
Multiple images of a person sequence were often used either to enhance
spatial feature descriptions of local image regions or patches
~\cite{gheissari2006person,farenzena2010person,cheng2011custom,xu2013human},
or to extract additional appearance information such as appearance change
statistics~\cite{bedagkar2012part}.
In contrast, the proposed model aims to simultaneously select and match
discriminative video appearance and space-time features for
maximising cross-view identity ranking. Our experiments show the
advantages of the proposed model over existing multi-shot models for
person ReID.

\section{Discriminative Video Ranking}
\label{sec:method}

\begin{figure*}[!t]
\centering
\includegraphics[width=1\linewidth]{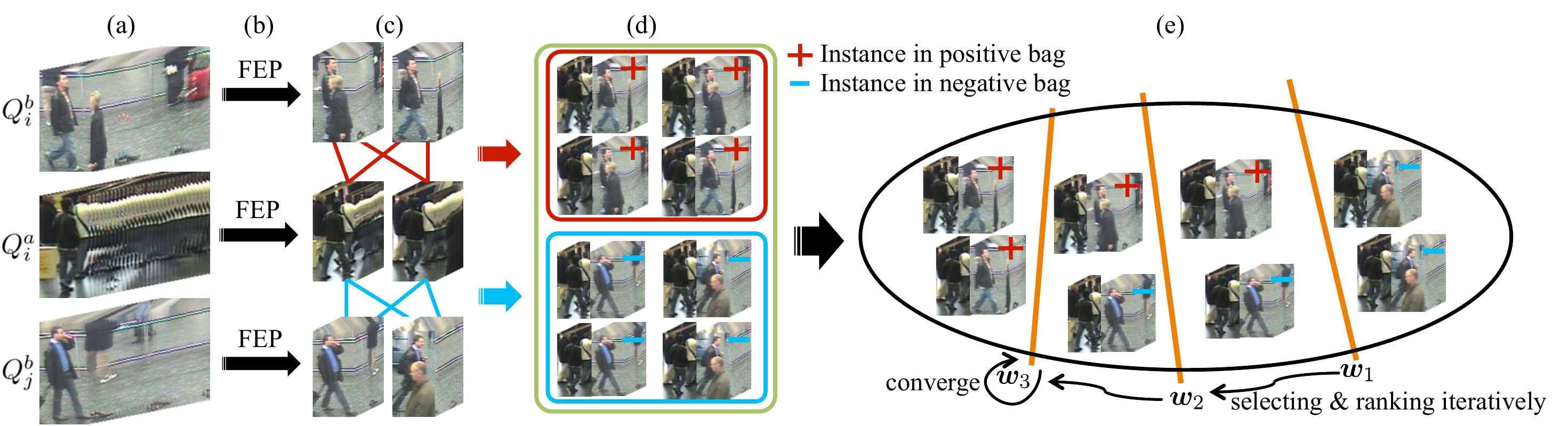}
\vskip -.3cm
\caption{\footnotesize
The training pipeline of the proposed discriminative video ranking framework.
(a) Training image sequences, $Q^a_i$ denotes the image sequence of person $p_i$ from camera $a$~(see Sec. \ref{sec:problem_definition}).
(b,c) Generating candidate fragments for each sequence 
(see Sec. \ref{sec:videofrag}).
(d) Creating cross-view fragment pairs as positive and negative instances and pooling them into positive and negative bags respectively~(see Sec. \ref{sec:sel_rank_ciseg} and Fig. \ref{fig:bag_formation}). 
(e) Learning a 
ranking model by simultaneously
selecting and ranking iteratively discriminative fragment
pairs~(see Sec. \ref{sec:sel_rank_ciseg}).
}
\label{fig:framework}
\vspace{-.6cm}
\end{figure*}

We formulate the person re-identification problem as a ranking problem~\cite{ProsserEtAlBMVC:10,Shaogang2014reid}.
Although image sequences of people may provide intuitively richer content
to learn discriminative information about an individual's visual appearance
when compared to a single still image widely used by existing person
ReID methods~\cite{li2013locally,gray2007evaluating,zheng2009associating,LoyCVPR09},
the availability of more (and often redundant) data poses additional challenges in model learning, 
e.g. more random inter-object occlusions and thus incomplete frames,
arbitrary sequence duration 
and uncertain
starting/ending postures,
\textcolor{black}{and potential clothing variations of some people over time}.
Moreover, human annotators may implicitly and unconsciously 
have the tendency to 
select carefully 
more clear and
better-segmented person images for learning image-based
ReID models. On the other hand, 
tracked sequences of person bounding boxes in 
typical surveillance videos are inherently more
noisy and incomplete. Directly utilising {\em all} the sequence data
for constructing ReID models can easily result in unstable
models, which is undesirable. A selection mechanism is
required to be 
part
of the learning method in order to optimally
explore the redundant information available in sequence data.

In the context of relative ranking based person ReID model
learning, it is non-trivial to automatically learn a robust
discriminative ranking function from such contaminated and uncontrolled image sequence
data. 
Inherently, one needs to address the problem of how to mitigate 
the negative influence of unknown noisy observations, e.g. various
types of occlusion and clutter in the background. This is beyond 
solving the more common problem of misalignment over time in sequence
matching.
In this work, we formulate a novel discriminative re-identification
model capable of simultaneously selecting and ranking informative
video fragments from pairs of unregulated person image sequences
captured in two non-overlapping camera views. 
%
Our model not only mitigates unwanted data whilst
exploring useful information from image sequences for person ReID, 
but also requires no
rigid sequence alignment as in the case of traditional methods, e.g. dynamic time warping.
%
Specifically, our model is based on\,: (i) Video fragmentation by
motion energy profiling~(Fig.~\ref{fig:framework}(b,c) and Sec.
\ref{sec:videofrag})\,; 
(ii) Learning a sequence based relative ranking function by simultaneously selecting and
ranking cross-view video fragment
pairs~(Fig. \ref{fig:framework}(d,e) and Sec.
\ref{sec:sel_rank_ciseg})\,. 
%
Once learned, 
our model can then be deployed to re-identify previously unseen people 
given cross-view 
unregulated image sequences~(Sec. \ref{sec:re_id_deployment}).
An overview diagram of 
the proposed approach
is presented in Fig.~\ref{fig:framework}.

\begin{figure*}
	\centering
	\includegraphics[width=0.9\linewidth]{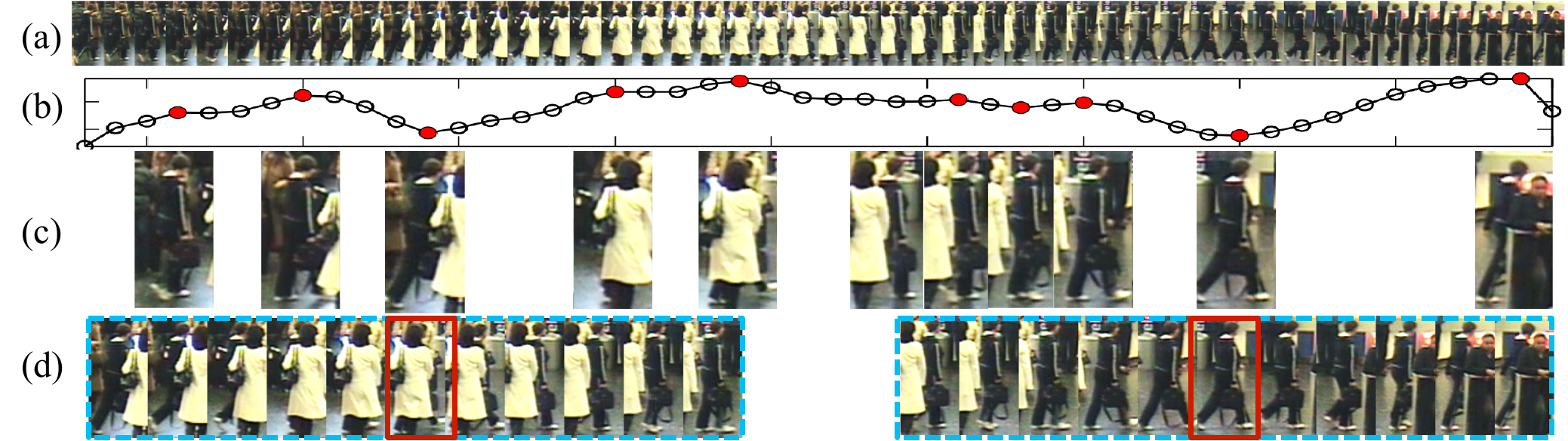}
	\vskip -.3cm
	\caption{\footnotesize
		(a) A person image sequence of $50$
		frames is shown, with the motion energy intensity for each frame given in (b).
		The red dots in (b) denote the automatically detected local minima and maxima
		temporal landmarks in the motion intensity profile, of which the
		corresponding frames are provided at the vertically-aligned positions in (c).
		(d) Two example video fragments (shown every $2$ frames)
		with the landmark frames highlighted by red bounding boxes.
	}
	\label{fig:flow_energy_profile}
	\vspace{-.6cm}
\end{figure*}

\subsection{Problem Definition}
\label{sec:problem_definition}

Suppose we have a collection of image sequence pairs 
$\{(Q_i^a, Q_i^b)\}_{i=1}^N$, 
where $Q_i^a$ and $Q_i^b$ denote the image sequences of person $p_i$ captured 
by two disjoint cameras $a$ and $b$,  
and $N$ the total number of training people.
Each image sequence $Q$ is defined 
by a set of consecutive frames $I$ as
$Q = (I_1, ..., I_{T})$, where $T$ is not a constant because in
typical surveillance videos,
tracked person image sequences \cite{Ben2011,Hare2011}
are not 
guaranteed to have
(1) a uniform duration (arbitrary frame numbers),
(2) the same number of walking cycles, 
(3) similar starting/ending postures, 
(4) high video frame rates, 
or (5) invariant clothing over time.

For model training, we aim to learn a ranking function $f(Q^a, Q^b)$ of image sequence pairs
that satisfies the following ranking constraints:
\begin{equation}
f(Q_i^a, Q_i^b) > f(Q_i^a, Q_j^b),\,\forall i=\{1,...,N\},\,\forall j\neq i,
\end{equation}
i.e. the sequence pair $(Q_i^a, Q_i^b)$ of the same person $p_i$ is constrained/optimised to 
have a higher rank over any cross-view sequence pairing 
of person $p_i$ and $p_j$ with $j \neq i$.

Learning a ranking function {\em holistically without discrimination
  and selection} from pairs of unsegmented and temporally unaligned
person image sequences will subject the learned model to significant
noise and degrade any meaningful discriminative information contained
in the image sequences. This is an inherent drawback of any holistic
sequence matching approach, including those with dynamic time warping
applied for non-linear mapping (see experiments in
Sec.~\ref{sec:experiments}). Reliable human parsing/pose
detection~\cite{kanaujia2007semi} or occlusion
detection~\cite{xiao2006bilateral} may help, but such approaches are
difficult to scale, especially with image sequences from crowded
public scenes.
The challenge is to
learn a robust ranking model effective in coping with
incomplete and partial image sequences by identifying and selecting
discriminative/informative video fragments from each sequence suitable for
extracting trustworthy fragment features. 
Let us first consider generating a pool of candidate fragments 
for each video, i.e. video fragmentation.

\subsection{Video Fragmentation}
\label{sec:videofrag}

Given unregulated image sequences of people, it is too noisy to
attempt to holistically locate and extract reliable discriminative
features from entire image sequences.  
Instead, we consider breaking down each sequence into a pool of 
localised video fragments 
to allow a learning model to
automatically select the discriminative fragments
(Sec.~\ref{sec:sel_rank_ciseg}).

It can be observed that motion energy intensity 
induced by the activity of human muscles during walking
exhibits regular periodicity \cite{waters1972electrical}. 
This motion energy intensity can be approximately estimated by optic flow computation.
We call this a Flow Energy Profile (FEP), see Fig.~\ref{fig:flow_energy_profile}.
This FEP signal is particularly suitable to address our 
video fragmentation problem
due to: (i) the local minima and
maxima landmarks probably correspond to characteristic
gestures of a walking process, 
and thus help in detecting them
(e.g. one foot is about to land); 
(ii) it is relatively robust to changes in camera viewpoint. 
More specifically, 
we first compute the optic flow field $(v_x, v_y)$ 
for each image frame $I$ from a sequence $Q$.  
Its flow energy 
is defined as 
\begin{equation}
e(I) = \sum_{(x,y)\in U}\lVert\left[\,v_x(x,y), v_y(x,y)\,\right]\rVert_2,
\label{eqn:flow_energy}
\end{equation}
where $U$ is the pixel set of the lower body, 
e.g. the lower half of $I$.
The FEP $\mathcal{E}$ of $Q$ is then obtained as
$\mathcal{E} = [e(I_1), ..., e(I_T)]$, which is further smoothed by a Gaussian filter to suppress noise.

Subsequently, we locate the local minima and maxima landmarks $\{\,t\,\}$ of $\mathcal{E}$ and for each landmark create a video fragment $s$ by extracting the surrounding frames $s = \{I_{t-L},...,I_{t},...,I_{t+L}\}$.
We fix $L=10$ for all our experiments, determined by cross-validation on the iLIDS-VID dataset.
Finally, we build a candidate set of video fragments $S = \{\,s\,\}$ by pooling all the fragments from $Q$.
Note that some fragments of each sequence can have similar walking phases since the local 
minima/maxima
landmarks of the FEP signal are likely to correspond to certain characteristic walking postures~(Fig. \ref{fig:flow_energy_profile}).
This increases the possibility of finding temporally aligned video fragment pairs
(i.e. centred at similar walking postures) given a pair of 
video fragment sets $(S^a, S^b)$ from two disjoint camera views,
facilitating discriminative video fragment selection and matching during model learning.
Also, Fig.~\ref{fig:flow_energy_profile} shows that 
the FEP signal can be sensitive to random occlusions and background
clutter that could lead to non-characteristic fragments.
However, this has limited impact on the overall effectiveness of the proposed selection-and-ranking model 
(Sec.~\ref{sec:sel_rank_ciseg}), as it is designed specifically to identify and 
exploit automatically discriminative video fragments
from largely redundant sets for 
training a ReID model.

\vspace{0.1cm}
\noindent \textbf{Video fragment representation}
To encode both the 
dynamic
and static appearance information of the subjects, 
we represent video fragments 
with both space-time and colour features.
%
They complement each other, especially in
the context of person ReID.
Colour features have been shown to be significant for
person
ReID~\cite{hirzer2012relaxed,Shaogang2014reid,LiuEtAl2012reid,liu2014fly,zhao2013person},
implicitly capturing the chromatic patterns of clothing independent from space-time characteristics of a person's appearance,
such as the way people walk. In contrast, the latter is encoded
by the space-time features. 

\vspace{0.1em}
\noindent \textit{Space-time feature} -- 
Particularly, we exploit HOG3D~\cite{klaser2008spatio} as the space-time feature representation of a video fragment, 
due to its advantages
demonstrated for applications in action and activity
recognition~\cite{wang2009evaluation,klaser2008spatio}. 
In order to capture spatially more detailed and localised space-time information of
a person in motion, we decompose 
a video fragment $s$ spatially into $2\times5$
uniform cells according to human biological 
{body} topology
such as head, torso, arms and legs.
To capture separately the information of sub-intervals before and after 
the characteristic walking posture (Fig.~\ref{fig:flow_energy_profile}\,(d)) 
potentially situated in the middle of a video fragment,
the fragment is further divided temporally into two smaller sub-phases,
\textcolor{black}{resulting in a total of $20$
  (i.e.\,$2$$\times$$5$$\times$$2$) cells for every video fragment.} 
Two adjacent cells have $50\%$ overlap for increased robustness to 
possible spatio-temporal fragment misalignment. 
A space-time gradient histogram is computed in each cell and then concatenated 
to form the HOG3D space-time descriptor $\bm{x}_\mathrm{st}$ of the fragment $s$.

\vspace{0.1em}
\noindent \textit{Colour feature} --
We adopt the localised average colour histogram
as the appearance feature of a video fragment
from a great number of alternative descriptors~\cite{Li20153542}
because of its simplicity and effectiveness~\cite{hirzer2012relaxed}.
Specifically, for each component frame in a video fragment, 
the colour features are extracted from rectangular patches 
($16$$\times$$8$ pixels in size) 
sampled from 
each frame
with an overlap of $8$ and $4$ pixels vertically and horizontally
between each patch
(i.e. $50\%$ overlap between adjacent patches).
In each patch, we compute the mean values of the HSV and LAB colour channels 
and form a framewise colour feature vector 
by concatenating the mean values of all the patches in a frame.
To minimise noise and obtain a more reliable colour representation,
all the framewise colour features of a fragment are averaged over time
to produce a fragment-wise appearance representation $\bm{x}_\mathrm{a}$ of that fragment $s$.

Finally, both space-time (HOG3D) and colour appearance (Colour) features $\bm{x}_\mathrm{st}$ and $\bm{x}_\mathrm{a}$ 
are concatenated into a fragment descriptor (ColHOG3D)
$\bm{x} = [\bm{x}_\mathrm{st};\bm{x}_\mathrm{a}]$.
Note, the image frames of all sequences are normalised into a fixed size
($128 \times 64$ pixels in our implementation) before computing any features.

\vspace{0.1cm}
\noindent \textit{Notations} -- 
Formally, for the $m$-th fragment $s^{a}_{i,m}$ from the person $p_i$'s image sequence captured in camera $a$, its descriptor is denoted by $\bm{x}^{a}_{i,m}$.
The same is for $s^{b}_{i,m}$ and $\bm{x}^{b}_{i,m}$.
%
We denote $X^a_i = \{\bm{x}^a_{i,m}\}_{m=1}^{|X^a_i|}$ and 
$X^b_i = \{\bm{x}^b_{i,m}\}_{m=1}^{|X^b_i|}$ as the descriptor set for
the fragments segmented from the sequences $Q_i^a$ and $Q_i^b$
of person $p_i$ in camera $a$ and $b$ respectively, where $|\cdot|$
represents the set cardinality. 
The entire collection of descriptors for $N$ training image sequence
pairs $\{(Q^a_i,Q^b_i)\}^N_{i=1}$
is denoted as $\{(X^a_i,X^b_i)\}^N_{i=1}$.

\subsection{Selection and Ranking}
\label{sec:sel_rank_ciseg}

As shown in Fig. \ref{fig:flow_energy_profile}, 
the fragments of a person image sequence can be contaminated by
unknown occlusions and background dynamics, and 
may also be extracted at an arbitrary time-instance of a walking
cycle.
Given such noisy fragment pair collections generated from cross-view image sequences, 
a significant challenge for sequence matching based ReID
is how to identify and select discriminative/informative and
temporally aligned fragment pairs (rather than the entire sequences)
to learn a suitable ranking model.
Formally, the objective is to learn a linear ranking function on 
the {entry-wise} absolute difference of two cross-view fragments $\bm{x}^a$ and $\bm{x}^b$:
\vspace{-.1cm}
\begin{equation}
h(\bm{x}^a, \bm{x}^b) 
= \bm{w}^\top \mathrm{abs}(\bm{x}^a - \bm{x}^b).
\label{eqn:ranking_function_original}
\vspace{-.1cm}
\end{equation}
We assume that for each person, there exists at least one cross-view fragment pair
that is sufficiently aligned over time and carries desired identity-sensitive information
for this person. Our aim is to construct a model capable of
automatically discovering and locating not only the best cross-view fragment pair
but also multiple cross-view fragment pairs that are sufficiently
aligned and discriminative for person ReID.
For model training with the best fragment pair,
it is equivalent to constraining a ranking function $h$ to 
prefer the most discriminative cross-view fragment pair of the same person $p_i$ 
to the pairings over $p_i$ and any other person $p_j$, $i \neq j$, \ie\,
\vspace{-.1cm}
\begin{equation}
( \max_{\bm{x}_{i,\cdot}^a \in X_i^a, \bm{x}_{i,\cdot}^b \in X_i^b} 
  h(\bm{x}_{i,\cdot}^a,\bm{x}_{i,\cdot}^b) ) 
> h(\bm{x}_{i,\cdot}^a,\bm{x}_{j,\cdot}^b), \,\forall\,j \ne i.
\label{eq:rank_constr}
\vspace{-.1cm}
\end{equation}

\begin{figure}[!t]
\centering
\includegraphics[width=1\linewidth]{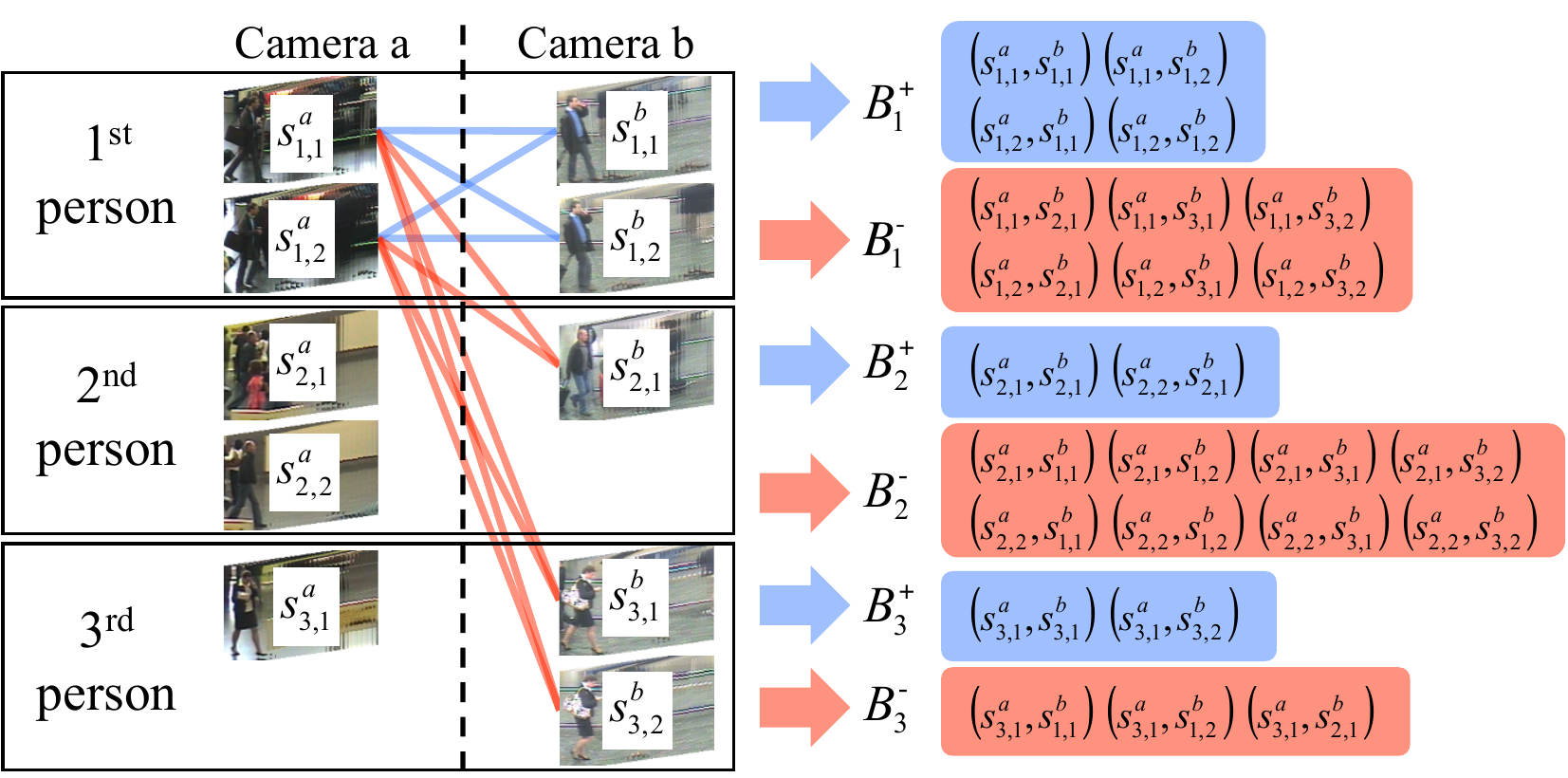}
\vskip -.3cm
\caption{\footnotesize
An overview of 
constructing the positive and negative
bags of inter-camera fragment-pairs. This is to generate the training
data from 
image sequences for DVR model
learning. Examples of three people are illustrated.
In particular, $s_{i,j}^\kappa$ denotes the $j$-th fragment from the $i$-th person image sequence captured
by camera $\kappa$, $\kappa \in \{a, b\}$. 
We build separately a positive ($B_i^+$) and negative ($B_i^-$) bag for the $i$-th person.
Consider the first person $p_1$ as an example,
the cross-view pairings (blue lines) 
on fragments from $p_1$ are used to form the positive bag $B_1^+$, 
whilst other pairings (red lines) between $p_1$ and a different person $p_i,
i\neq1$, shall create the negative bag $B_1^-$.
}
\label{fig:bag_formation}
\vspace{-.5cm}
\end{figure}%
For notation simplicity, we define $\bm{y}_{i,m}^+ = \mathrm{abs}(\bm{x}^a_{i,\cdot} - \bm{x}^b_{i,\cdot})$ as the $m$-th \textit{positive instance} of person $p_i$, i.e., the {entry-wise absolute difference} of two cross-view fragments of the same person $p_i$, and $\bm{y}_{i,m}^- = \mathrm{abs}(\bm{x}^a_{i,\cdot} - \bm{x}^b_{j,\cdot}), j\neq i$ as the $m$-th \textit{negative instance}, i.e., the absolute difference of two cross-view fragments of $p_i$ and another person.
For each person $p_i$, we form a \textit{positive bag} $B^+_i = \{\bm{y}_{i,m}^+\}_{m=1}^{|B^+_i|}$ by pooling the positive instances, and a \textit{negative bag} $B^-_i = \{\bm{y}_{i,m}^-\}_{m=1}^{|B^-_i|}$ by pooling the negative instances.
The formation process of 
positive and negative bags for individual persons is illustrated in Fig. \ref{fig:bag_formation}.
Note, we only consider a single directional pairing (from camera $a$
to $b$) without considering the opposite direction
when constructing negative bags. This is because our empirical
experiments suggest that the addition of negative instances from
camera $b$ to $a$
only gains negligible ($<$$0.3\%$) ReID performance advantage
whilst the additional cost is significantly more ($>$$200\%$) and
complex in both bag construction and model learning.
A plausible explanation is that the negative instances from camera
$a$ to $b$ are sufficiently diverse (in our experiments) and
bi-directional negative sampling does not add meaningfully richer data. 
This is also supported by that only $10\%$ of the full negative
instances from camera $a$ to $b$ were utilised and shown to be
sufficient in model learning, with the added benefit of reducing
the number of pairwise constraints (the first inequality constraint in Eqn.~(\ref{eq:mirank}))
in model learning, therefore speeding up the training process.

By redefining the ranking function 
$h(\bm{x}^a, \bm{x}^b) = g(\mathrm{abs}(\bm{x}^a - \bm{x}^b)) = g(\bm{y})$,
Eqn.~(\ref{eq:rank_constr}) can be rewritten as
\vspace{-.1cm}
\begin{equation}
( \max_{\bm{y}_{i,\cdot}^+ \in B^+_i} g(\bm{y}_{i,\cdot}^+) ) 
       > g(\bm{y}_{i,\cdot}^-), \forall\,\bm{y}_{i,\cdot}^-\in B^-_i.
\label{eqn:ranking_function_simple}
\vspace{-.1cm}
\end{equation}
With the ranking constraints in Eqn.~(\ref{eqn:ranking_function_simple}),
we aim to automatically discover and select the most discriminative/informative and temporally aligned cross-view fragment pair $\bm{y}_{i,\cdot}^+$
within the positive bag $B_i^+$ for each person $p_i$ 
for learning an identity discriminative ranking model. 
To that end,
we introduce a binary selection 
variable $\bm{v}_i$
with each entry being either $0$ or $1$ and of unity $\ell_0$ norm
for each person $p_i$, and then obtain
\vspace{-.1cm}
\begin{equation}
g(\bm{Y}_i\bm{v}_i) > g(\bm{y}_{i,\cdot}^-), \forall\,\bm{y}_{i,\cdot}^-\in B^-_i,
\label{eq:ranking_function_simple_sel}
\vspace{-.1cm}
\end{equation}
where each column of $\bm{Y}_i$ corresponds to one $\bm{y}^+ \in B^+_i$, $||\bm{v}_i||_0 = 1,\; \bm{e}^\top\bm{v}_i=1$
\textcolor{black}{, and $\bm{e}$ denotes a vector of all ``1''s.}

To achieve good generalisation ability for the ranking model given the
ranking constraints in
Eqn.~(\ref{eq:ranking_function_simple_sel}), we formulate our
problem as a max-margin ranking problem by defining the
objective function as: 
\vspace{-.1cm}
\begin{equation}
\label{eq:mirank}
\begin{split}
&\bm{w}^*=\arg\min_{\bm{w}, \bm{v}, \bm{\xi}}\frac{1}{2}||\bm{w}||^2 + C\bm{e}^\top \bm{\xi}\\
&\text{s.t. } \; \bm{v}_i^\top \bm{Y}_i^\top \bm{w} - (\bm{y}_{i,m}^-)^\top \bm{w} \ge 1 - \xi_{i,m},\;\\  
&\quad \;\;\;\; \xi_{i,m} \ge 0,\; \forall\,\bm{y}_{i,m}^- \in B^-_i,\; m \in \{1,\dots, |B^-_i|\},\\
&\quad \;\;\; ||\bm{v}_i||_0 = 1,\; \bm{e}^\top\bm{v}_i=1, 
\;\; i \in \{1,\dots,N\}.\\
\end{split}
\vspace{-.2cm}
\end{equation}
where \textcolor{black}{$\bm{w}$ is the parameter of the objective ranking function defined in Eqn. (\ref{eqn:ranking_function_original})}, 
and $N$ the number of people in the training set.
$\bm{v}$ is the concatenation of the binary selection variables of all persons: 
$\bm{v} = [\bm{v}_1; \bm{v}_2; ...\,\bm{v}_N]$.
$\bm{\xi}$ is the flattened slack variable, formed by all the possible $\xi_{i,m}$.
We solve Eqn. (\ref{eq:mirank}) by iteratively optimising $\bm{w}$ and $\bm{v}$ between a ranking step and a selecting step.

\vspace{0.1em}
\noindent\textbf{Ranking step} 
We fix $\bm{v}$ to optimise $\bm{w}$.
Eqn. (\ref{eq:mirank}) turns into
\vspace{-.1cm}
\begin{equation}
\label{eq:ranking_step}
\begin{split}
&\bm{w}^*=\arg\min_{\bm{w}, \bm{\xi}}\frac{1}{2}||\bm{w}||^2 + C\bm{e}^\top \bm{\xi}\\
&\text{s.t. } \; \bm{v}_i^\top \bm{Y}_i^\top \bm{w} - (\bm{y}_{i,m}^-)^\top \bm{w} \ge 1 - \xi_{i,m},\;\\  
&\quad \;\;\;\; \xi_{i,m} \ge 0,\; \forall\,\bm{y}_{i,m}^- \in B^-_i,\; m \in \{1,\dots, |B^-_i|\},\\
&\quad \;\;\;\; i \in \{1,\dots,N\}.\\
\end{split}
\vspace{-.1cm}
\end{equation}
With the fragment selections $\bm{v}$ known, Eqn. (\ref{eq:ranking_step}) is a standard RankSVM problem and can be efficiently solved with a primal training algorithm~\cite{chapelle2010efficient}. 

\vspace{0.1em}
\noindent\textbf{Selecting step}
We fix $\bm{w}$ to optimize $\bm{v}$.
The term on $\bm{w}$ (i.e. $\frac{1}{2}||\bm{w}||^2$)
can be eliminated and Eqn. (\ref{eq:mirank}) becomes
\vspace{-.1cm}
\begin{equation}
\label{eq:selecting_step}
\begin{split}
&\bm{v}^*=\arg\min_{\bm{v}, \bm{\xi}}\bm{e}^\top \bm{\xi}\\
&\text{s.t. } \; \bm{v}_i^\top \bm{Y}_i^\top \bm{w} - (\bm{y}_{i,m}^-)^\top \bm{w} \ge 1 - \xi_{i,m},\\
&\quad \;\;\;\; \xi_{i,m} \ge 0,\; \forall\,\bm{y}_{i,m}^- \in B^-_i,\; m \in \{1,\dots, |B^-_i|\}\\
&\quad \;\;\;\; ||\bm{v}_i||_0 = 1,\; \bm{e}^\top\bm{v}_i=1, 
\;\; i \in \{1,\dots,N\}.\\
\end{split}
\vspace{-.1cm}
\end{equation}
Considering that the person-wise $\bm{v}_i$ is associated only with 
$\{\xi_{i,m}\}^{|B^-_i|}_{m = 1}$ and we are optimising the summation of all possible $\xi_{i,m}$, Eqn.\,(\ref{eq:selecting_step}) is equivalent to optimising $\bm{v}_i$ 
for each person $p_i$ separately, as
\vspace{-.1cm}
\begin{equation}
\label{eq:selecting_step_person}
\begin{split}
&\bm{v}_i^*=\arg\min_{\bm{v}_i, \bm{\xi}_i}\bm{e}^\top \bm{\xi}_i\\
&\text{s.t. } \; \bm{v}_i^\top \bm{Y}_i^\top \bm{w} - (\bm{y}_{i,m}^-)^\top \bm{w} \ge 1 - \xi_{i,m},\\
&\quad \;\;\;\; \xi_{i,m} \ge 0,\;\forall\,\bm{y}_{i,m}^- \in B^-_i,\; m \in \{1,\dots, |B^-_i|\}\\
&\quad \;\;\;\; ||\bm{v}_i||_0 = 1,\; \bm{e}^\top\bm{v}_i=1.\\
\end{split}
\vspace{-.2cm}
\end{equation}
where $\bm{\xi}_i = [\xi_{i,1}, \dots, \xi_{i,|B^-_i|}]^\top$.
The inequality constraints in Eqn. (\ref{eq:selecting_step_person}) 
can be transformed as
\vspace{-.1cm}
\begin{equation}
\begin{split}
&\xi_{i,m} \ge 1 - \bm{v}_i^\top \bm{Y}_i^\top \bm{w} + (\bm{y}_{i,m}^-)^\top \bm{w}, \;\;
\text{s.t.} \;\; \xi_{i,m} \ge 0.
\end{split}
\vspace{-.1cm}
\end{equation}
Therefore, for any particular $\bm{v}_i \in V$ 
that holds $||\bm{v}_i||_0 = 1$ and $\bm{e}^\top\bm{v}_i=1$ in the selecting space $V$, 
the entries $\xi^*_{i,m}$ of the optimal $\bm{\xi}^*_i$ that minimises the summation $\bm{e}^\top\bm{\xi}_i$ shall be
\vspace{-.1cm}
\begin{equation}
\xi^*_{i,m} = \max\{0, 1 - \bm{v}_i^\top \bm{Y}_i^\top \bm{w} + (\bm{y}_{i,m}^-)^\top \bm{w}\}.
\label{eqn:xi_star}
\vspace{-.1cm}
\end{equation}
It is obvious that the summation $\bm{e}^\top\bm{\xi}_i$ is a function of $\bm{v}_i$,
\vspace{-.2cm}
\begin{equation}
\begin{split}
&q(\bm{v}_i)=\sum_{m=1}^{|B^-_i|} {\xi^*_{i,m}} \\
&\quad \quad \, =\sum_{m=1}^{|B^-_i|} {\max\{0, 1 - \bm{v}_i^\top \bm{Y}_i^\top \bm{w} + (\bm{y}_{i,m}^-)^\top \bm{w}\}}.
\end{split}
\vspace{-.2cm}
\end{equation}
Finally we can obtain the $\bm{v}^*_i$ by optimising $q(\bm{v}_i)$ via:
\vspace{-.1cm}
\begin{equation} \footnotesize
\label{eq:optimize_v}
\begin{split}
&\bm{v}_i^* = \arg\min_{\bm{v}_i \in V}q(\bm{v}_i)\\
&\,\,\,\,\,\, =\arg\min_{\bm{v}_i \in V}\sum_{m=1}^{|B^-_i|} {\max\{0, 1 - \bm{v}_i^\top \bm{Y}_i^\top \bm{w} + (\bm{y}_{i,m}^-)^\top \bm{w}\}},\\
&\,\,\,\,\,\,\,\, \text{s.t. } \; ||\bm{v}_i||_0 = 1,\; \bm{e}^\top\bm{v}_i=1.
\end{split}
\vspace{-.1cm}
\end{equation}
For each person $p_i$, we only have a limited number of $\bm{v}_i$ in
$V$. Therefore Eqn. (\ref{eq:optimize_v}) can be efficiently solved 
even with a greedy search.

To begin the model training process, we set $\bm{v}_i =
\frac{1}{|B_i^+|} \bm{e}$ to initiate a balanced/moderate start 
since the quality of $\bm{y}_{i,\cdot}^+$ is unknown {\em a priori}.
The iteration terminates when $\bm{v}_i$ 
does not change any more.
Typically, the training process stops after {$4\sim5$} iterations.
For learning efficiency, $10\%$ out of all the $\bm{y}^-_{i,\cdot}$ 
are randomly selected to form $B_i^-$.
Since only a single $\bm{y}_{i,\cdot}^+$ for each person $p_i$ is selected and 
utilised for model learning, we call this model \textbf{DVR(single)}.

\subsubsection{Multiple Cross-View Fragment Pair Selection}
\label{sec:method_multiple_selection}
Thus far we have detailed the procedure of training our DVR(single) model via identifying the \textit{best} cross-view fragment pair in each positive bag $B^+_i$ (corresponding to person $p_i$) for learning the ranking function (Eqn.~(\ref{eqn:ranking_function_original})).
This allows us to largely avoid the contamination effect from harmful data.
Nonetheless, we may simultaneously lose some useful information 
from discarding the majority {of} instances $\bm{y}_{i,\cdot}^+$ of each bag $B^+_i$,
because some of these ignored $\bm{y}_{i,\cdot}^+$ can be of good quality.
Identifying and exploiting these
``good though not the best''
fragment data $\bm{y}_{i,\cdot}^+$ is likely to
benefit the model learning.
To that end, we shall describe next our multiple cross-view fragment pair selection algorithm for 
better exploring image sequence data.

Our multiple fragment-pair selection algorithm is 
based on a goodness/quality measure of 
individual $\bm{y}_{i,\cdot}^+$.
Once all instances $\bm{y}_{i,\cdot}^+$ of person $p_i$ 
are measured by assigning a score $\gamma_{i,\cdot}$ 
(higher is better) to each instance, 
we can easily locate multiple (top $k$) discriminative $\bm{y}_{i,\cdot}^+$ 
from the ranked list of all $\bm{y}_{i,\cdot}^+$ sorted in descending order 
of $\gamma_{i,\cdot}$.
Formally, we define $\gamma_{i,\cdot}$ 
for each $\bm{y}_{i,\cdot}^+$ as
\vspace{-.1cm}
\begin{equation}
\gamma_{i,\cdot} = \sum_{m=1}^{|B_i^-|} (1 - \xi_{i,m}^*).
\label{eq:quality_measure}
\vspace{-.1cm}
\end{equation}
We denote $1 - \xi_{i,m}^*$ as the ranking margin of $\bm{y}_{i,\cdot}^+$
against 
$\bm{y}_{i,m}^-$, which can be obtained by
Eqn.~(\ref{eqn:xi_star}).
Given Eqn.~(\ref{eq:quality_measure}), the $\bm{y}_{i,\cdot}^+$ with a larger 
cumulated ranking margin
over all the negative instance $\bm{y}_{i,m}^-$ is preferred. 
This formulation generalises
the single selection case that searches for the best $\bm{v}_i^*$ 
(Eqn.~(\ref{eq:optimize_v})), 
i.e. the $\bm{v}_i^*$ and the highest $\gamma_{i,\cdot}$ 
leads to
the same selection of positive instance $\bm{y}_{i,\cdot}^+$.

After the top $k$ $\bm{y}_{i,\cdot}^+$ for each person $p_i$ are found and selected,
we can obtain multiple (i.e. $k$) $\bm{v}_i^*$s 
by setting the corresponding entry of each $\bm{v}_i^*$ to ``$1$''
whilst the remaining entries to ``$0$''.
We call this model \textbf{DVR(top$\bm{k}$)}.
Similar to the single selection model DVR(single), 
these ranking constraints associated with the selected top $k$ $\bm{y}_{i,\cdot}^+$
are then employed
for optimising $\bm{w}$ with Eqn.~(\ref{eq:ranking_step}).
In Sec.~\ref{sec:exp_model_analysis}, we shall 
evaluate the effect of different 
top $k$ positive instances on the person ReID performance.
An overview of learning the proposed DVR model is presented in Algorithm~\ref{Alg:DVR}.

\begin{algorithm} [t] \scriptsize
	\caption{\footnotesize DVR Model Learning}
	\label{Alg:DVR}
	\SetAlgoLined
	\KwIn
	{ 
		Training image sequence pairs $\{(Q^a_i,Q^b_i)\}^N_{i=1}$;
	}	 
	\KwOut
	{
	  	The ranking function $\bm{w}$ (Eqn.~(\ref{eqn:ranking_function_original}));
	}
 	%
	
	\textbf{(I) Video fragmentation} (Sec.~\ref{sec:videofrag}): \\
	- Segment each $Q$ into a set of fragments $\{s\}$;\\
	- Extract space-time and appearance features $\bm{x}$ from $s$;
	
	\textbf{(II) Bag construction} (Fig.~\ref{fig:bag_formation}): for each person $p_i$, \\
	- Form a positive bag $B_i^+$ with positive instance $\bm{y}_{i,\cdot}^+$; \\
	- Form a negative bag $B_i^-$ with negative instance $\bm{y}_{i,\cdot}^-$; \\
	
	\textbf{(III) Model Learning} (Sec.~\ref{sec:sel_rank_ciseg}): \\
	\textit{/* Initialise selection vectors */}: \\
	{		
		$\bm{v_i} = \frac{1}{|B_i^+|}$, $i=1, ..., N$; \\
	}
	\While{$\mathrm{true}$}
	{		
		\textit{/* Ranking step */}: \\
		Obtain $\bm{w}^*$ with fixed $\{\bm{v}_i\}$ (Eqn.~(\ref{eq:ranking_step})); \\
		
		\textit{/* Selecting step */}: \\
		\For{$i=1, ..., N$}
 		{
 			\If{$\mathrm{single\; selection}$}
 			{
 				Obtain $\bm{v}_i^*$ (Eqn.~(\ref{eq:optimize_v}));
 			}
 			\Else
 			{
 				\textit{/* Multiple selection */}: \\
 				Compute $\gamma_{i,\cdot}$ for each $\bm{y}_{i,\cdot}^+$ (Eqn.~(\ref{eq:quality_measure}));\\
 				Rank $\bm{y}_{i,\cdot}^+$ in descendant order of $\gamma_{i,\cdot}$; \\
 				Find the top-$k$ $\bm{y}_{i,\cdot}^+$; \\
 				Obtain $k$ $\bm{v}_i^*$s (Sec.~\ref{sec:method_multiple_selection});
 			}
			 			
 		}
 		
 		\textit{/* Convergence check */}: \\
 		\If{$\mathrm{no}$ $\bm{v}_i$ $\mathrm{changed}$}
 		{
 			Return $\bm{w}^*$.
 		}
 	} 
\end{algorithm}

\subsubsection{Model Complexity}
\label{sec:model_complexity}
We analyse the training complexity of the DVR model, 
focusing on the ranking and selecting steps.
For model training, 
we adopt the primal RankSVM scheme~\cite{chapelle2010efficient} as the ranking solver.
Its complexity is $O(cd^2) + O(d^3)$ due to Hessian computation and
the linear search in Newton direction respectively, 
with $c$ and $d$ denoting the number of ranking constraints 
(see Equations~(\ref{eq:rank_constr}) and (\ref{eq:ranking_step})) 
and the feature dimensions.
Suppose $k$ positive instances per person are selected in the training stage,
then $c = k\sum_{i=1}^N |B_i^-|$,
where $N$ is the total number of training people.

The cost for the selection process mainly involves measuring the quality score 
of each positive instance of all 
training people with Eqn.~(\ref{eqn:xi_star}) and Eqn.~(\ref{eq:quality_measure}).
Its complexity is $O(c d u)$, where $u = \sum_{i=1}^N |B_i^+|$ 
denotes the total number of positive instances across all training data.
The total complexity of model training 
is thus $O(cd^2 + d^3 + cdu )$.
We evaluated and reported the model training cost in our
experiments (Sec.~\ref{sec:exp_model_analysis}).

\subsection{Re-Identification by DVR}
\label{sec:re_id_deployment}

Once learned, the ranking model (Eqn.~(\ref{eqn:ranking_function_original})) 
can be deployed to perform person re-identification by
matching a given probe person image sequence $Q^p$ observed in one camera
view against a gallery set $\{Q^g\}$ in another disjoint camera.
Formally, the ranking/matching score of a gallery person sequence $Q^g$
with respect to $Q^p$ is computed as
\vspace{-.1cm}
\begin{equation}
f({Q^p}, {Q^g}) = \max_{\bm{x}_{i,\cdot} \in X^p, \bm{x}_{j,\cdot} \in X^g}
\bm{w}^\top \mathrm{abs}(\bm{x}_{i,\cdot} - \bm{x}_{j,\cdot}),
\label{eqn:ranking_score_deployment}
\vspace{-.1cm}
\end{equation}
where $X^p$ and $X^g$ are the 
feature sets of the video fragments 
extracted from the sequences $Q^p$ and $Q^g$, respectively.
The same video fragmentation process as used for model training
(Sec.~\ref{sec:videofrag}) is employed for deploying a trained model. 
Finally, the gallery people are sorted in descending order of their assigned matching
scores to generate a ranking list.

\vspace{0.1cm}
\noindent \textbf{Combination with prior spatial feature based models} 
Our approach can complement existing spatial feature based person re-identification
approaches.
In particular, we incorporate Eqn.~(\ref{eqn:ranking_score_deployment}) 
into the ranking scores $\mathcal{R}_i$ obtained by other models as
\vspace{-.1cm}
\begin{equation}
\hat{f}({Q^p}, {Q^g}) = \sum_i \alpha_i \mathcal{R}_i({Q^p}, {Q^g}) + f({Q^p}, {Q^g}),
\label{eqn:combination}
\vspace{-.1cm}
\end{equation}
where $\alpha_i$ refers to the weighting assigned to the $i$-th method,
which is estimated by cross-validation.

\begin{figure*} 
	\centering
	\subfigure 
	{
		\includegraphics[width=0.38\linewidth]{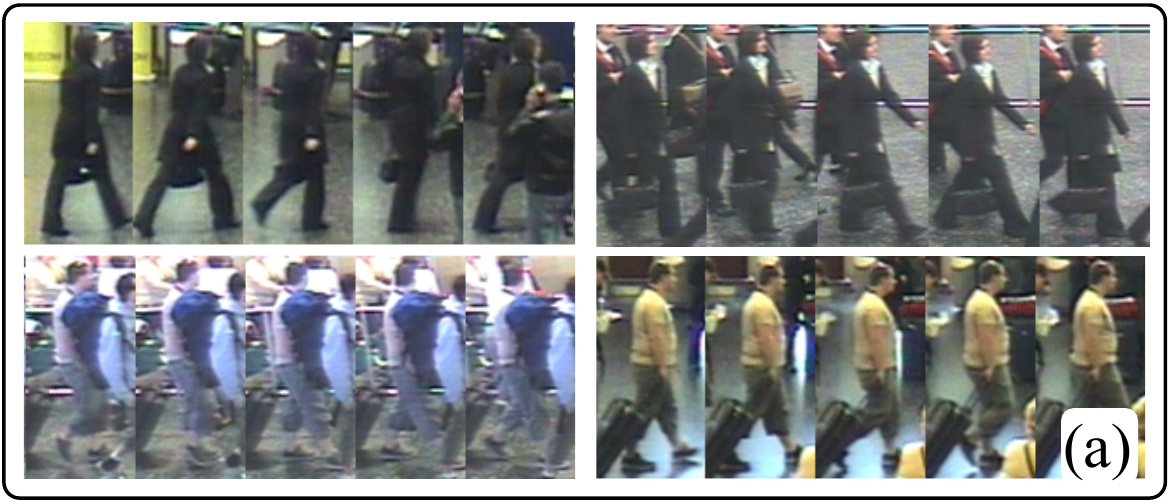}
	}
	\subfigure 
	{
		\includegraphics[width=0.38\linewidth]{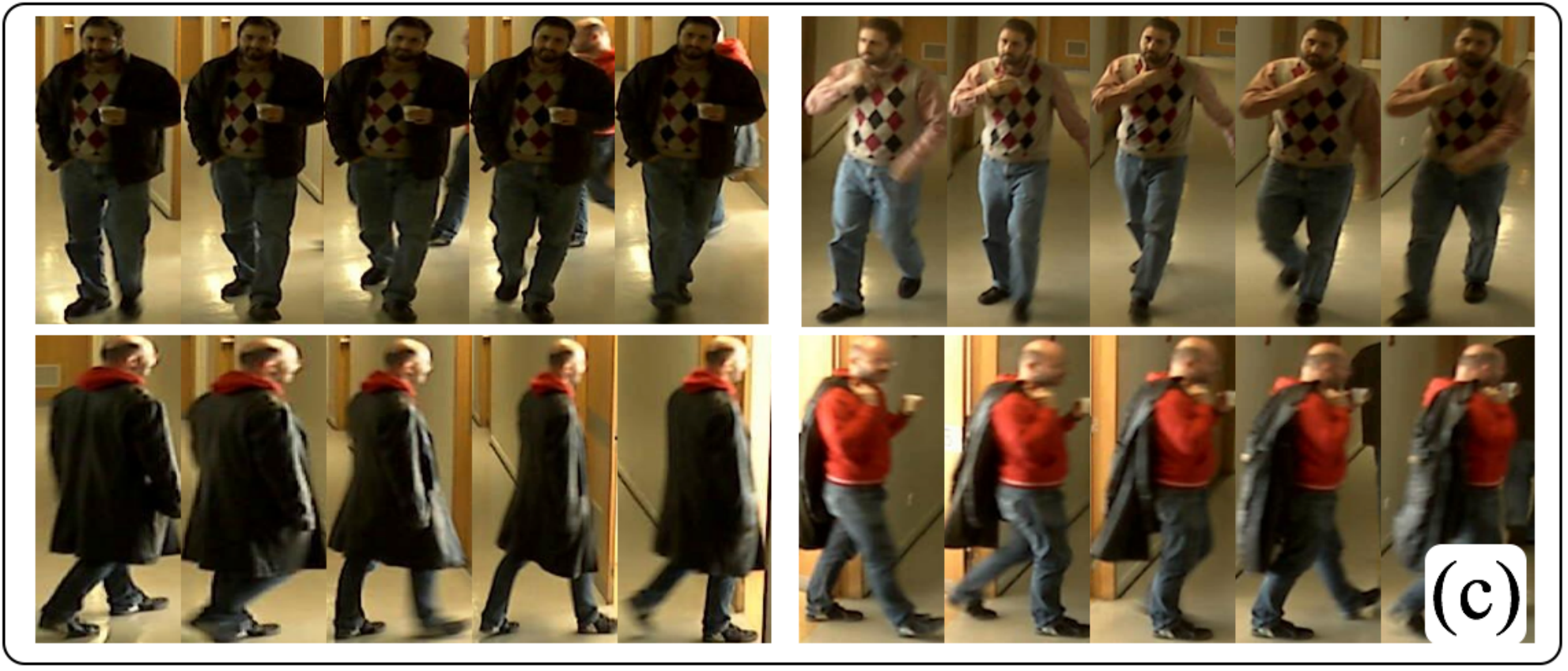}
	} 	
	\\ [-.15cm]
	\subfigure 
	{
		\includegraphics[width=0.38\linewidth]{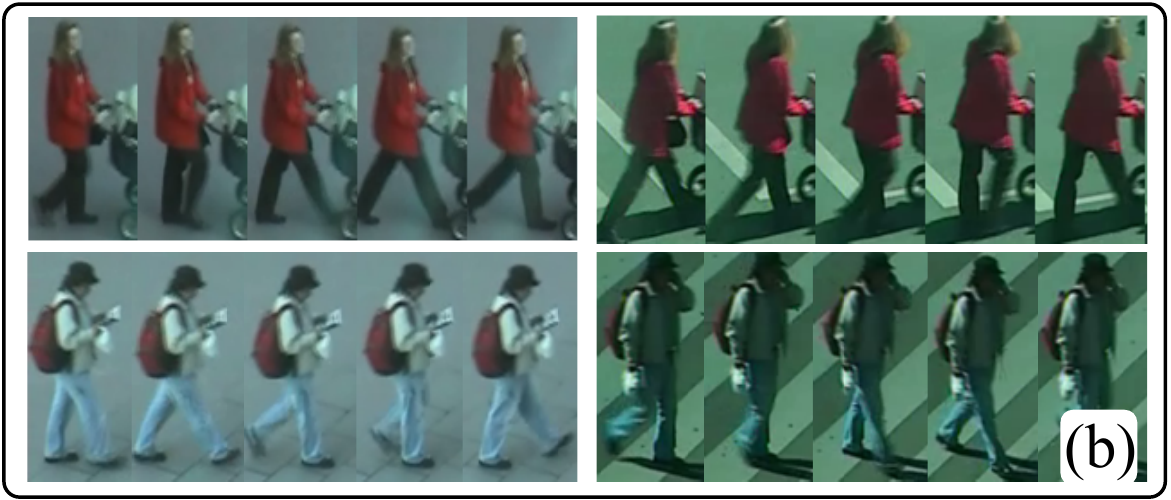}
	}
	\subfigure 
	{
		\includegraphics[width=0.38\linewidth]{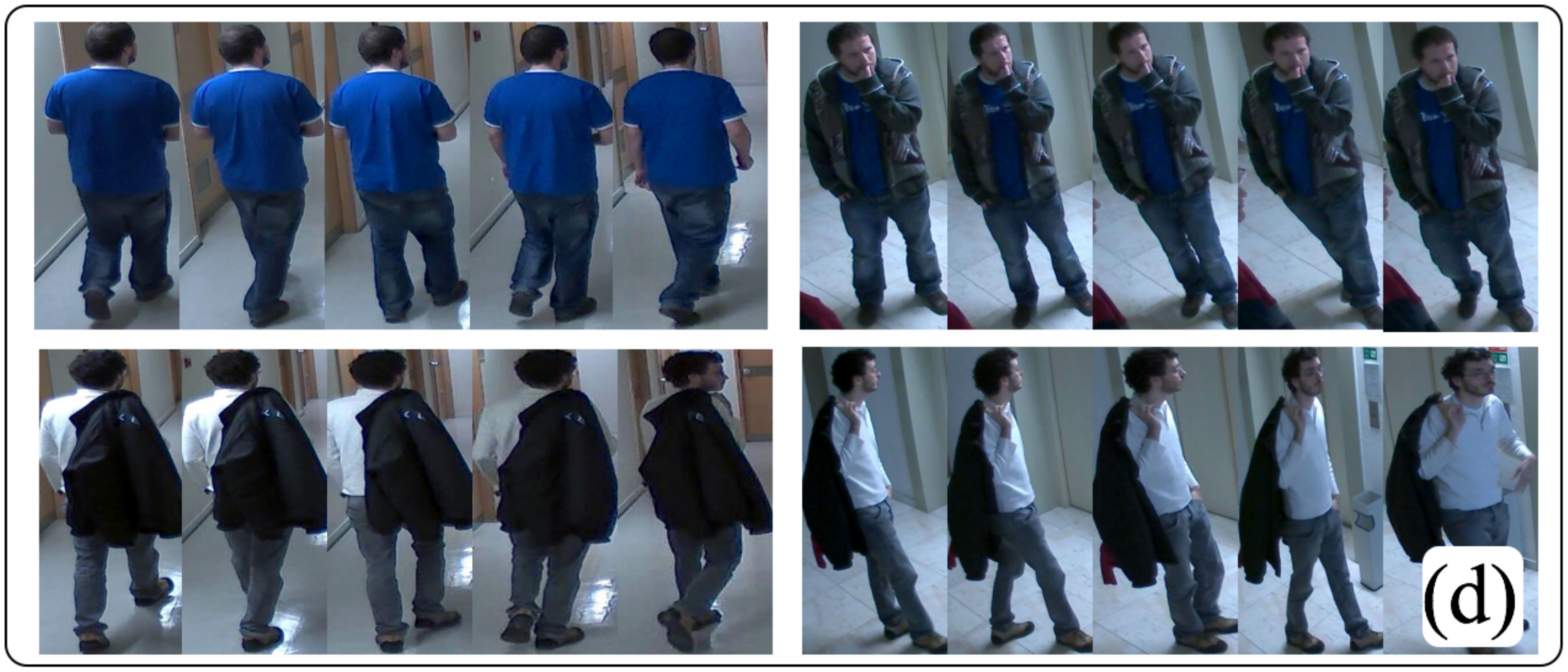}
	}
	\vskip -.3cm
	\caption{\footnotesize
		\textcolor{black}{
			Example pairs of image sequences from the same people
			appearing in different camera views given by
			(a) iLIDS-VID~\cite{TaiqingECCV14},
			(b) PRID$2011$~\cite{hirzer11a},
			(c) a $5$fps (frames per second) camera pair $(19, 40)$ from HDA+~\cite{figueira2014hda}, and
			(d) a $2$fps camera pair $(50, 57)$ from HDA+~\cite{figueira2014hda}.
			Note, severe appearance changes occur in
                        HDA+ due to explicit clothing variation for
                        some people in HDA+ ((c,d)); whilst
                        significant
                        appearance changes also occur
                        in iLIDS-VID
                        due to severe occlusion (bottom pair of (a)),
                        and mostly less extreme appearance changes exist in
                        iLIDS-VID and PRID$2011$ due to changes in
                        viewpoint and lighting ((a,b)).
		}
	}
	\label{fig:dataset_example}
	\vspace{-.6cm}
\end{figure*}

\subsection{Discussions on Related Models}
\label{sec:discussion}
We discuss the relationship of our proposed DVR model with 
other relevant contemporary models in the literature, with a focus on their differences.
First, most existing max-margin ranking methods~\cite{chapelle2010efficient,ProsserEtAlBMVC:10}
do not consider uncertainty in the ranking constraints during model optimisation.
In contrast, the proposed DVR model jointly optimises both the
selection of the ranking constraints and the ranking function. This is
necessary because the bag-level (e.g. image sequences) supervision 
cannot directly determine the instance-level (e.g. fragments) constraints
(Sec.~\ref{sec:sel_rank_ciseg}).

Second,
our model also differs notably from other multi-instance ranking models~\cite{Bergeron2008,bergeron2012fast,hu2008multiple}
in a number of aspects.
(1) Bergeron et al.~\cite{Bergeron2008} relaxed the selection vectors $\bm{v}_i$ 
(Eqn.~(\ref{eq:ranking_function_simple_sel}))
to be continuous during model optimisation, 
whilst our model searches for exact solutions of instance selection. 
As shown in our evaluation (Sec.~\ref{sec:exp_model_analysis}), 
Bergeron et al.'s relaxation method can significantly increase the
cost of constraint selection when the training set is large, though it does not compromise the model performance. 
(2) The model presented in~\cite{hu2008multiple} 
focuses on encoding bag-level (or sample-level) constraints into the ranking function 
by modelling instance-level constraints, assuming all instances can provide contribution
to model optimisation.
In contrast, we emphasise the selection of discriminative/informative instance data
(e.g. fragments) for robust learning, necessary for coping with very noisy and incomplete data 
(e.g. unregulated image sequences), whilst the stronger assumption
made in \cite{hu2008multiple} is less valid.
(3) Different from all these multi-instance models~\cite{Bergeron2008,bergeron2012fast,hu2008multiple},
the proposed DVR model is unique in 
its capability for allowing different quantities 
of explicit discriminative instance
selection and then exploitation, due to our formulation
of a principled instance quality measure (Eqn.~(\ref{eq:quality_measure})).
This can potentially increase the flexibility and scalability of our model 
in a variety of problem settings (e.g. varying degrees of noise) 
and applications (e.g. other sequence matching based tasks).

\section{Experiments}
\label{sec:experiments}

\noindent \textbf{Datasets}
Extensive experiments were conducted 
on three image sequence datasets
designed for person ReID,  
iLIDS Video re-IDentification (iLIDS-VID)~\cite{TaiqingECCV14}, 
PRID$2011$~\cite{hirzer11a}, and
HDA+ \cite{figueira2014hda}.
All three datasets are very challenging 
due to
clothing similarities among people, 
lighting and viewpoint variations
across camera views, cluttered background and occlusions
(Fig.~\ref{fig:ambiguous_appearance} and Fig.~\ref{fig:dataset_example}).

\vspace{0.1cm}
\noindent\textit{iLIDS-VID} --
Our new iLIDS-VID person sequence dataset~\cite{TaiqingECCV14} was created based on two
non-overlapping camera views from the i-LIDS Multiple-Camera
Tracking Scenario (MCTS)~\cite{ilids_mctc08}, which was captured at an airport arrival
hall under a multi-camera CCTV network (Fig.~\ref{fig:dataset_example}(a)). 
It consists of $600$ image sequences for $300$ randomly sampled
people, with one pair of image sequences from two disjoint camera views for
each person. Each image sequence has a variable length consisting of
$23$ to $192$ image frames, with an average number of $73$.

\vspace{0.1cm}
\noindent\textit{PRID$2011$} --
The PRID$2011$ dataset~\cite{hirzer11a} includes $400$
image sequences for $200$ people from two camera views that are adjacent to each other (Fig.~\ref{fig:dataset_example}(b)).
Each image sequence has a variable length consisting of
$5$ to $675$ image frames\footnote{We used sequences of $>$$21$ frames from $178$ people in the evaluation. 
	}, with an average number of $100$. 
Compared with the iLIDS-VID dataset, 
it is less challenging due to being captured in non-crowded
outdoor scenes with relatively simple and clean backgrounds and rare occlusions.

\vspace{0.1cm}
\noindent\textit{HDA+} --
The HDA+ dataset \cite{figueira2014hda} 
contains a total of $83$ labelled people
across $13$ indoor cameras in an office environment
(Fig.~\ref{fig:dataset_example}(c,d)).
HDA+ is characterised by
(i) 
low and variable frame rates, e.g. $2$$\sim$$5$fps (frames per second)
of HDA+ 
{\em versus} $25$fps of both PRID$2011$ and iLIDS-VID; and
(ii) 
clothing variation over time.
One limitation of HDA+ is the small number of people re-appearing
between camera pairs whilst re-appearance is required for evaluating ReID.
In our experiments, we selected two camera pairs, 
$(19, 40)$ and $(50, 57)$, that satisfy:
(1) a sufficiently large number of people reappearing across the camera views;
(2) very low video frame rates to evaluate its effect on space-time feature based ReID models;
(3) some people's clothing changes to evaluate the clothing-variation challenge. 
In particular, camera pair $(19, 40)$
provides
pairwise image sequences of $28$ different people at $5$fps. Each video has
$15$$\sim$$227$ frames with an average of $88$ frames.
In contrast, camera pair $(50, 57)$ contains pairwise videos of $10$
people at only $2$fps, with sequence length varying
between
$1$$\sim$$136$ frames
with
an average of $31$ frames.
For sequences $<$$21$ frames, we expanded them up to $21$ frames by
interpolating new frames using duplicates of the temporally-nearest frames in a sequence.
This is to enable fragmentation on them.
Note, little or no space-time information is available in very short sequences, 
e.g. 
$1$ frame. This is designed to test how a space-time
feature based model degrades with decreasing space-time information
available in the input video data.

\begin{table*}[!t] \scriptsize
	\renewcommand{\arraystretch}{1.15}
	\centering
	\caption{\footnotesize
		Compare different DVR variants.
		Fragments are represented by HOG3D \& colour. 
		RT: Ranking Time; 
		ST: Selecting Time;
		Unit is second.
	}
	\vskip -0.3cm
	\label{tab:model_selection}
	\begin{tabular}
		{p{1.3cm}|
			p{.2cm}p{.2cm}p{.2cm}p{.3cm}|p{.06cm}p{.2cm}|
			p{.2cm}p{.2cm}p{.2cm}p{.3cm}|p{.06cm}p{.2cm}|
			p{.2cm}p{.2cm}p{.2cm}p{.3cm}|p{.1cm}p{.24cm}|
			p{.2cm}p{.2cm}p{.2cm}p{.3cm}|p{.1cm}p{.24cm}
		}		
		\hline
		{Dataset} & 
		\multicolumn{6}{c|}{PRID$2011$\cite{hirzer11a}} & 
		\multicolumn{6}{c|}{iLIDS-VID\cite{TaiqingECCV14}} &
		\multicolumn{6}{c|}{\textcolor{black}{HDA+($5$fps)\cite{figueira2014hda}}} & 
		\multicolumn{6}{c}{\textcolor{black}{HDA+($2$fps)\cite{figueira2014hda}}} \\ 
		\hline
		Rank $R$ \textcolor{black}{(\%)}
		& 1 & 5 & 10 & 20 & RT & ST
		& 1 & 5 & 10 & 20 & RT & ST
		& \textcolor{black}{1} & \textcolor{black}{2} & \textcolor{black}{3} & \textcolor{black}{4} & \textcolor{black}{RT} & \textcolor{black}{ST}
		& \textcolor{black}{1} & \textcolor{black}{2} & \textcolor{black}{3}  & \textcolor{black}{4}  & \textcolor{black}{RT} & \textcolor{black}{ST} \\
		\hline
		\hline
		DVR(float)
		&{38.9}	&{68.8}	&{81.1}	&{91.3} & \textbf{6} & \textbf{8} 
		&{36.8}	&{59.3}	&{70.9}	&{80.1} & \textbf{28} & 740  
		&\textcolor{black}{54.3}  &\textcolor{black}{70.0}  &\textcolor{black}{77.9}  &\textcolor{black}{85.0}  &\textcolor{black}{2.2}  &\textcolor{black}{0.09} 
		&\textcolor{black}{52.0}  & \textcolor{black}{70.0} &\textcolor{black}{90.0}  &\textcolor{black}{100}  &\textcolor{black}{2.0}  &\textcolor{black}{0.06} 
		\\
		\hline
		DVR(single)
		&{38.9}	&{68.8}	&{81.1}	&{91.3} & \textbf{6} & 9 
		&{36.8}	&{59.3}	&{70.9}	&{80.1} & \textbf{28} & \textbf{97} 
		&\textcolor{black}{ 54.3} & \textcolor{black}{ 70.0} & \textcolor{black}{ 77.9} & \textcolor{black}{ 85.0} &\textcolor{black}{2.2}  &\textcolor{black}{\bf 0.06} 
		&\textcolor{black}{ 52.0} & \textcolor{black}{ 70.0} & \textcolor{black}{ 90.0} & \textcolor{black}{ 100} &\textcolor{black}{2.0}  &\textcolor{black}{\bf 0.01} 
		\\
		DVR(top$2$)
		& 39.4 & 70.6 & 83.7 & 91.8 & 13 & 9 
		& 37.7 & 60.1 & 71.1 & 81.4 & 42 & \textbf{97}  
		& - & - & - & - & - & -  
		& - & - & - & - & - & -  
		\\
		DVR(top$3$)
		& 40.0 & \textbf{71.7} & 84.5 & 92.2 & 15 & 9 
		& 39.5 & 61.1 &	71.7 & 81.0 & 58 & \textbf{97} 
		&-  &-  &-  &-  &- &-
		& - & - & - & - & - & - 
		\\
		DVR(top$4$)
		& 40.0 & 71.6 &	84.0 & 92.8 & 20 & 9 
		& 39.2 & \textbf{62.3} &71.7 & \textbf{81.9} & 70 & \textbf{97}
		& - & - & - & - & - & -
		& - & - & - & - & - & -	
		\\
		DVR(top$5$)
		& \textbf{40.8} & \textbf{71.7} & \textbf{84.9} & \textbf{93.1} & 21 & 9 
		& \textbf{39.9} & 62.1 &\textbf{71.9} & \textbf{81.9} & 81 & \textbf{97}  
		& - & - & - & - & - & -
		& - & - & - & - & - & - 
		\\
		\hline
	\end{tabular}
	\vspace{-0.2cm}
\end{table*}

\begin{table*}[!t] \scriptsize
	\renewcommand{\arraystretch}{1.15}
	\centering
	\caption{\footnotesize
		Compare different video fragment representations
		using the DVR(single) model. }
	\vskip -0.3cm
	\label{tab:model_feature}
	\begin{tabular}
		{l|
			p{.2cm}p{.2cm}p{.2cm}p{.4cm}|
			p{.2cm}p{.2cm}p{.2cm}p{.4cm}|
			p{.2cm}p{.2cm}p{.2cm}p{.4cm}|
			p{.2cm}p{.2cm}p{.2cm}p{.4cm}
		}	
		\hline
		Dataset & 
		\multicolumn{4}{c|}{PRID$2011$\cite{hirzer11a}} & 
		\multicolumn{4}{c|}{iLIDS-VID\cite{TaiqingECCV14}} &
		\multicolumn{4}{c|}{\textcolor{black}{HDA+($5$fps)\cite{figueira2014hda}}} & 
		\multicolumn{4}{c}{\textcolor{black}{HDA+($2$fps)\cite{figueira2014hda}}} 
		\\
		\hline
		Rank $R$ \textcolor{black}{(\%)} 
		& 1 & 5 & 10 & 20
		& 1 & 5 & 10 & 20
		& \textcolor{black}{1} & \textcolor{black}{2} & \textcolor{black}{3} & \textcolor{black}{4}
		& \textcolor{black}{1} & \textcolor{black}{2} & \textcolor{black}{3}  & \textcolor{black}{4} \\
		\hline
		\hline
		HOG3D
		&{28.9}	&{55.3}	&{65.5}	&{82.8} 
		&{23.3}	&{42.4}	&{55.3}	&{68.4}
		& \textcolor{black}{40.0} & \textcolor{black}{52.1} & \textcolor{black}{65.0} & \textcolor{black}{70.0}
		& \textcolor{black}{18.0} & \textcolor{black}{30.0} & \textcolor{black}{62.0} & \textcolor{black}{76.0}   
		\\
		\hline
		\textcolor{black}{Colour}
		&\textcolor{black}{30.1}&\textcolor{black}{54.3}&\textcolor{black}{64.9}&\textcolor{black}{79.7} 
		&\textcolor{black}{24.2}&\textcolor{black}{44.6}&\textcolor{black}{56.0}&\textcolor{black}{67.4}  
		&\textcolor{black}{48.6}&\textcolor{black}{62.1}&\textcolor{black}{71.4}&\textcolor{black}{81.4}
		&\textcolor{black}{40.0}&\textcolor{black}{64.0}&\textcolor{black}{\bf 90.0}&\textcolor{black}{\bf 100}
		\\
		\hline
		ColHOG3D
		&\textbf{38.9}&\textbf{68.8}&\textbf{81.1}&\textbf{91.3}
		&\textbf{36.8}&\textbf{59.3}&\textbf{70.9}&\textbf{80.1}
		&\textcolor{black}{\bf 54.3} & \textcolor{black}{\bf 70.0} & \textcolor{black}{\bf 77.9} & \textcolor{black}{\bf 85.0} 
		&\textcolor{black}{\bf 52.0} & \textcolor{black}{\bf 70.0} & \textcolor{black}{\bf 90.0} & \textcolor{black}{\bf 100} 
		\\
		\hline
	\end{tabular}
	\vspace{-0.5cm}
\end{table*}

\vspace{0.1cm}
\noindent \textbf{Evaluation settings}
From every dataset, 
all sequence pairs are randomly split
into two subsets of equal size, 
one for training and one for
testing.
Following the evaluation protocol on the PRID$2011$ dataset~\cite{hirzer11a}, in the testing phase, 
the sequences from one camera are used as the probe set while the ones from another camera are the gallery set.
%
	The results are measured by Cumulated Matching Characteristics (CMC).
	Specifically, we show top rank matching rates. As CMC values are
        proportional to the dataset size (the overall population for the
        ranked pairs), we adopt Ranks $1$$\sim$$20$ for PRID$2011$ and
        iLIDS-VID, and Ranks $1$$\sim$$4$ for HDA+
        ($<$$1/5$ size of iLIDS-VID and PRID$2011$), 
        so that these
        values are approximately comparable across all four datasets.
%
To obtain stable statistical results, we repeat the experiments for $10$ trials and report the average results.

\subsection{Evaluation on Model Variants}
\label{sec:exp_model_analysis}

We evaluated and analysed the proposed DVR model in 
three aspects:
(1) effectiveness of the selection mechanisms;
(2) effectiveness of the fragment representations;
(3) robustness against low and variable video frame rates.

\vspace{0.1cm}
\noindent \textcolor{black}{\em Effectiveness of the selection mechanisms} --
For the selection mechanism, we conducted two comparisons: 
(a) the DVR(single) model \textit{versus} our preliminary model
reported in~\cite{TaiqingECCV14}
which we call \textbf{DVR(float)} since 
its selection involves a (float) weighted combination of instances
in contrast to our new single or multiple explicit instance selection
strategies,
(b) single \textit{versus} multiple 
fragment-pair selection
(Sec.~\ref{sec:method}).
The results in Table~\ref{tab:model_selection} (the first two rows)
show that identical scores
are obtained by DVR(single) and DVR(float)~\cite{TaiqingECCV14}.
This is further verified by the observation that both models select
almost identical discriminative video fragments. 
On the other hand, the computational cost/time required are
different for the two models, in particular when the visual content 
is more crowded and selection becomes harder. More
specifically, for model training including both the ranking and
selecting steps, Table~\ref{tab:model_selection} shows that both models
require similar time for the ranking step 
on all datasets. This is because they are subject to the same number of
ranking constraints (Eqn.~(\ref{eq:ranking_step})).
However, although the time required for the selection routine is
similar for PRID$2011$ \textcolor{black}{and HDA+}, DVR(single) is significantly
faster than DVR(float) on iLIDS-VID, 
e.g. over~$7\times$ speed up.
This was performed on a $64$-bit Intel CPU Processor@$2.7$GHz with
a MATLAB implementation in Linux OS. 
These observations suggest 
no advantage in treating the selection
as a float weighted combination of instances 
as originally proposed in~\cite{TaiqingECCV14,Bergeron2008}.

One may ask the question how many discriminative fragment pairs should
be selected from each cross-view image sequence pair of a person 
during model training.
To that end, we evaluated the performance of ReID
using different numbers of positive fragment pairs per person
\textcolor{black}{on PRID$2011$ and iLIDS-VID}\footnote{
	\textcolor{black}{
		This multi-fragment selection evaluation is not performed on HDA+ as
		some short image sequences have only one fragment.
		}
	}.
It is evident from Table~\ref{tab:model_selection} that
the use of additional discriminative fragment pairs 
can further boost the overall performance of person ReID 
at the price of increased model training time. 
This empirically supports our analysis on the potential benefits of
multiple fragment pair selection and exploitation as discussed in
Sec.~\ref{sec:method_multiple_selection}. 
However, the margin of improvement from additional fragment data quickly
diminishes. In our experiments, we utilised up to the top $5$ fragment pairs per
person. Any further addition of more pairs had very limited effect in
improving the learned ranking model. Moreover, it is
also observed that the construction of ranking constraints in RankSVM
is a time consuming process and its complexity is linear in the number
of constraints. Empirically, selecting the top
$3$ discriminative fragment pairs from a matched training image sequence pair
for model learning provides a good trade-off between 
ReID accuracy and model learning cost. 
For the remaining experiments reported in this section, 
DVR(top$3$) models were
trained for PRID$2011$ and iLIDS-VID 
and DVR(single) models for HDA+
in the comparative
evaluation against other baseline methods.

\begin{table*}[!t] \scriptsize
	\renewcommand{\arraystretch}{1.15}
	\caption{\footnotesize
		Comparison with gait recognition and temporal sequence matching methods.
	}
	\vskip -0.3cm
	\label{tab:compare_alternatives}
	\centering
	\begin{tabular}
		{p{2.9cm}|
			p{.2cm}p{.2cm}p{.2cm}p{.4cm}|
			p{.2cm}p{.2cm}p{.2cm}p{.4cm}|
			p{.2cm}p{.2cm}p{.2cm}p{.4cm}|
			p{.2cm}p{.2cm}p{.2cm}p{.4cm}
		}
		\hline
		Dataset & 
		\multicolumn{4}{c|}{PRID$2011$\cite{hirzer11a}} & 
		\multicolumn{4}{c|}{iLIDS-VID\cite{TaiqingECCV14}} &
		\multicolumn{4}{c|}{\textcolor{black}{HDA+($5$fps)\cite{figueira2014hda}}} & 
		\multicolumn{4}{c}{\textcolor{black}{HDA+($2$fps)\cite{figueira2014hda}}} \\
		\hline
		Rank $R$ \textcolor{black}{(\%)}
		& 1 & 5 & 10 & 20 
		& 1 & 5 & 10 & 20
		& \textcolor{black}{1} &\textcolor{black}{2} & \textcolor{black}{3} & \textcolor{black}{4} 
		& \textcolor{black}{1} & \textcolor{black}{2} & \textcolor{black}{3}  & \textcolor{black}{4}  \\
		\hline
		\hline
		\textcolor{black}{Gait Recognition\cite{martin2012gait}}
		&	\textcolor{black}{20.9}	&	\textcolor{black}{45.5}	&	\textcolor{black}{58.3}	&	\textcolor{black}{70.9}	
		&	2.8		&	13.1	&	21.3	&	34.5
		&	\textcolor{black}{34.3}		&	\textcolor{black}{47.9}	&	\textcolor{black}{57.9}&	\textcolor{black}{62.9}
		&	\textcolor{black}{24.0}		&	\textcolor{black}{40.0}	&	\textcolor{black}{62.0}	&	\textcolor{black}{78.0} \\
		\hline
		ColLBP\cite{hirzer2012relaxed}+DTW\cite{rabiner1993fundamentals}    
		&	14.6	&	33.0	&	42.6	&	47.8
		&	9.3	   	&	21.7	&	29.5	&	43.0
		&	\textcolor{black}{40.7}		&	\textcolor{black}{55.7}&	\textcolor{black}{62.9}&	\textcolor{black}{74.3}
		&	\textcolor{black}{40.0}		&	\textcolor{black}{66.0}&	\textcolor{black}{76.0}	&	\textcolor{black}{86.0}\\
		HoGHoF\cite{laptev2008learning}+DTW\cite{rabiner1993fundamentals}  	    
		&	17.2	&	37.2	&	47.4	&	60.0
		&	5.3	    &	16.1	&	29.7	&	44.7
		&	\textcolor{black}{11.4}		&	\textcolor{black}{22.1}&	\textcolor{black}{32.9}&	\textcolor{black}{48.6}
		&	\textcolor{black}{36.0}		&	\textcolor{black}{46.0}&	\textcolor{black}{58.0}&	\textcolor{black}{84.0} \\
		ColLBPHoGHoF+DTW\cite{rabiner1993fundamentals}  	    
		&	14.7	&	33.5	&	42.7	&	47.8
		&	10.1    &	22.5	&	29.9	&	43.6
		&	\textcolor{black}{44.3}		&	\textcolor{black}{57.9}&	\textcolor{black}{62.9}	&	\textcolor{black}{74.3}
		&	\textcolor{black}{40.0}		&	\textcolor{black}{66.0}&	\textcolor{black}{74.0}&	\textcolor{black}{76.0} \\
		\hline
		\textbf{DVR} 
		&\textbf{40.0} &\textbf{71.7} &\textbf{84.5} &\textbf{92.2}
		&\textbf{39.5} &\textbf{61.1} &\textbf{71.7} &\textbf{81.0}
		&\textcolor{black}{\bf 54.3} & \textcolor{black}{\bf 70.0} & \textcolor{black}{\bf 77.9} & \textcolor{black}{\bf 85.0} 
		&\textcolor{black}{\bf 52.0} & \textcolor{black}{\bf 70.0} & \textcolor{black}{\bf 90.0} & \textcolor{black}{\bf 100}  
		\\
		\hline
	\end{tabular}
	\vspace{-.5cm}
\end{table*}

\begin{figure}[!t]
	\centering
	\subfigure
	{
		\includegraphics[width=0.9\linewidth]{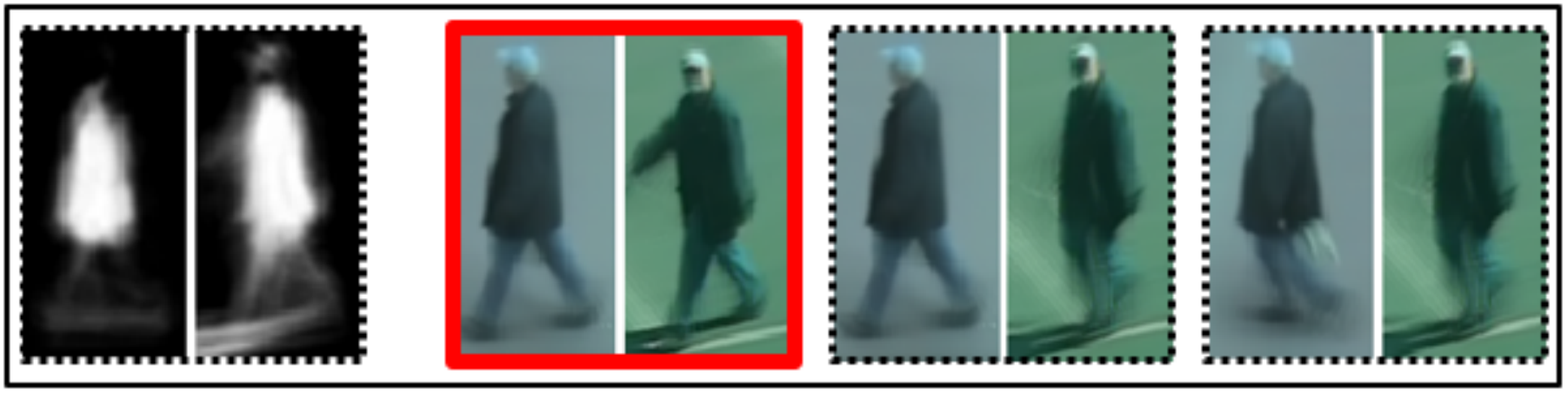}%
	}
	\vskip -.15cm
	\subfigure
	{
		\includegraphics[width=0.9\linewidth]{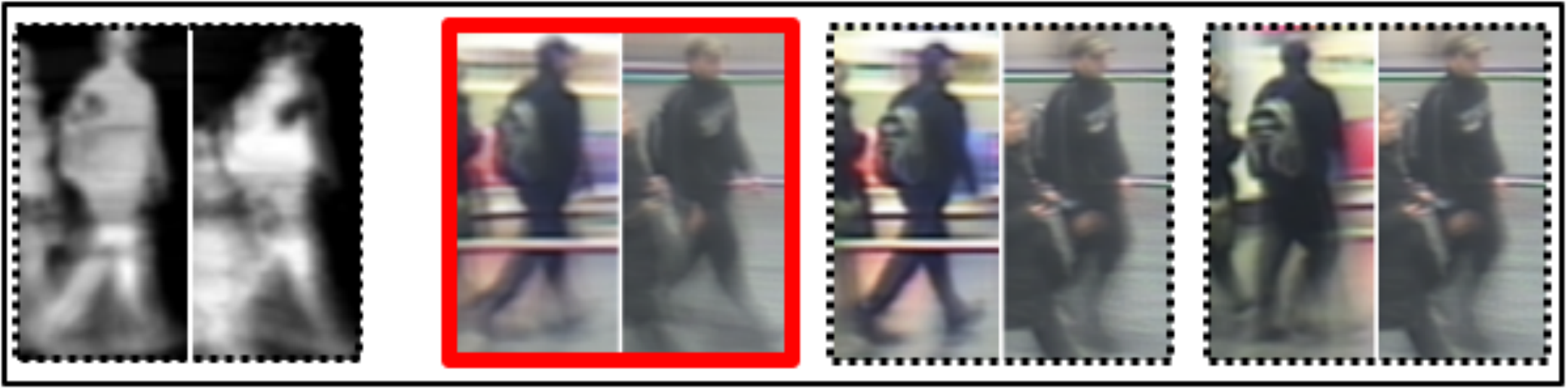}%
	}
	\vskip -.15cm
	\subfigure
	{
		\includegraphics[width=0.9\linewidth]{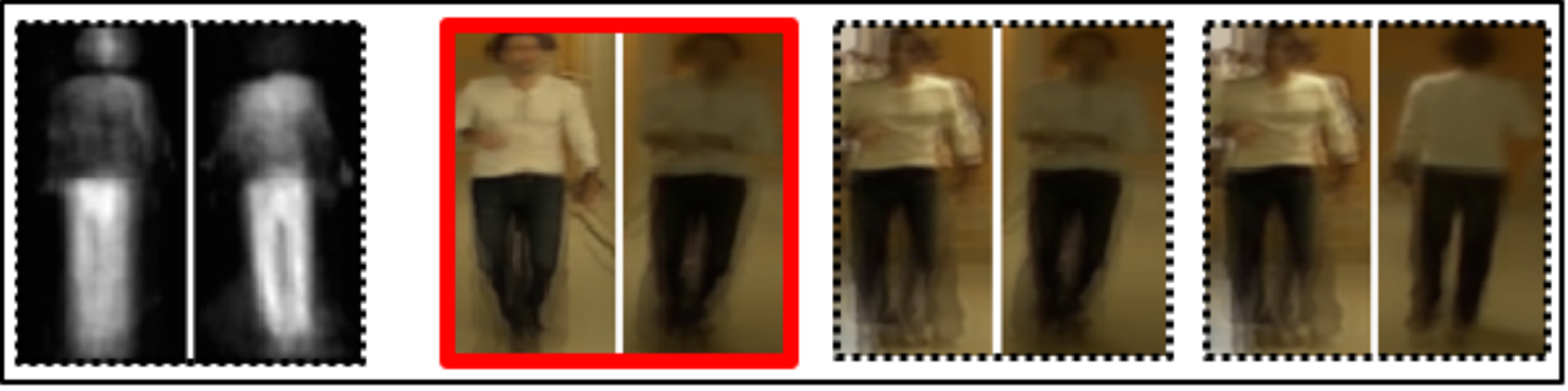}%
	}
	\vskip -.3cm
	\caption{\footnotesize
		Three examples of the GEI gait features 
		and the DVR video fragment pairs.
		Top: from PRID$2011$; Middle: from iLIDS-VID; Bottom: from HDA+($5$fps). 
		In each example, the leftmost thumbnail shows GEI gait features, while the remaining thumbnails present some examples of fragment pairs, with the automatically selected pairs marked by red bounding boxes.
		A fragment is visualised as the weighted average of all its frames with emphasis on its central frame.
	}
	\label{fig:gei}
	\vspace{-.6cm}
\end{figure}

\vspace{0.1cm}
\noindent \textcolor{black}{\em Effectiveness of the fragment representations} --
It is worth pointing out that our preliminary work presented
in~\cite{TaiqingECCV14} is somewhat limited on fragment representation
as no colour appearance information is considered. 
Here we report a significant improvement in performance from combining
the space-time features (HOG3D) with  
colour features (Sec. \ref{sec:videofrag}). 
\textcolor{black}{
For the DVR(single) model, 
Table~\ref{tab:model_feature} shows
$34.6\%$, $57.9\%$, $35.8\%$ and $188.9\%$ increase at Rank-1 recognition rate 
on PRID$2011$, iLIDS-VID, HDA+($5$fps) and HDA+($2$fps) respectively, 
when comparing with the results by HOG3D and ColHOG3D.
This suggests that colour plays an important role in re-identifying
people, also evident from the colour-only ReID performance in the table.}
These results demonstrate the importance of utilising both
space-time and colour appearance information for person
ReID in image sequence data, further supporting previous
studies on the importance of leveraging colour information 
for ReID 
\cite{hirzer2012relaxed,Shaogang2014reid,LiuEtAl2012reid,liu2014fly,zhao2013person}.
Throughout the following experiments, ColHOG3D is adopted as the default fragment representation in our DVR model,
unless specified otherwise.

\vspace{0.1cm}
\noindent
	{\em Robustness against low and variable
    video frame-rates} --  
	The proposed DVR model is expected to benefit more from higher
        frame-rate videos, 
	whilst its advantage over appearance-only based models
        diminishes gradually with a decrease in 
        video frame rate as less space-time information is available. 
	The results in Table \ref{tab:model_feature} show that
        the space-time feature (HOG3D) {\em only} based DVR model
        produces very competitive ReID accuracy compared to models using
        colour features alone, given high ($25$fps) frame rate
        videos from PRID$2011$ and iLIDS-VID.  
	Encouragingly, HOG3D-only based DVR retains credible ReID
        accuracies on $5$fps sequences from HDA+. However, when the
        frame rate decreases more significantly to $2$fps, the
        performance of the HOG3D-only based model degrades considerably
	whilst the colour-only based DVR is less affected.
	These results are consistent with the expectation that
        space-time feature alone based ReID models degrade when very
        limited or no space-time information is available in very 
        low frame rate videos. 
	Nevertheless, the space-time information selected by the DVR
        model is still useful for ReID even at such a low frame rate.
	It is also evident that the full DVR model using the ColHOG3D
        representation selectively explores the 
        complementary information from both space-time and  
	colour appearance features for 
	significant improvements
	on ReID accuracies in all situations including very low video
        frame-rates (the bottom row in Table~\ref{tab:model_feature}).
	This illustrates the strength and robustness of the DVR model
        in utilising complementary visual information,  
	even when space-time information is very poor or even absent. 
	This also demonstrates the robustness and flexibility of the
        DVR model in coping with significant variations in video
        frame rate when extracting and exploiting discriminative
        space-time information from unregulated surveillance videos.

\subsection{Comparing Gait Recognition and Temporal Sequence Matching}
\label{sec:gait_dtw}


\begin{table*}[!t] \scriptsize
	\renewcommand{\arraystretch}{1.15}
	\caption{\footnotesize
		Comparing spatial appearance feature based ReID methods. 
		SS: Single-Shot; MS: Multi-Shot.
	}
	\vskip -0.3cm
	\label{tab:compare_stateart}
	\centering
	\begin{tabular} 
		{l|
			p{.2cm}p{.2cm}p{.2cm}p{.4cm}|
			p{.2cm}p{.2cm}p{.2cm}p{.4cm}|
			p{.2cm}p{.2cm}p{.2cm}p{.4cm}|
			p{.2cm}p{.2cm}p{.2cm}p{.4cm}
		}
		\hline
		Dataset & 
		\multicolumn{4}{c|}{PRID$2011$\cite{hirzer11a}} & 
		\multicolumn{4}{c|}{iLIDS-VID\cite{TaiqingECCV14}} &
		\multicolumn{4}{c|}{\textcolor{black}{HDA+($5$fps)\cite{figueira2014hda}}} & 
		\multicolumn{4}{c}{\textcolor{black}{HDA+($2$fps)\cite{figueira2014hda}}}
		\\
		\hline
		Rank $R$ \textcolor{black}{(\%)}
		& 1 & 5 & 10 & 20
		& 1 & 5 & 10 & 20 
		& \textcolor{black}{1} & \textcolor{black}{2} & \textcolor{black}{3} & \textcolor{black}{4}
		& \textcolor{black}{1} & \textcolor{black}{2} & \textcolor{black}{3}  & \textcolor{black}{4}
		\\
		%
		\hline \hline
		SS-ColLBP\cite{hirzer2012relaxed}
		&	22.4	&	41.8	&	51.0	&	64.7
		&	9.1	    &	22.6	&	33.2	&	45.5
		&	\textcolor{black}{35.7}		&	\textcolor{black}{47.1}&	\textcolor{black}{55.0}	&	\textcolor{black}{58.6}
		&	\textcolor{black}{16.0}	&	\textcolor{black}{52.0}&	\textcolor{black}{74.0}&	\textcolor{black}{90.0} \\
		\hline
		SS-SDALF\,\cite{farenzena2010person}	
		&	4.9	&	21.5	&	30.9	&	45.2
		&	5.1	&	14.9	&	20.7	&	31.3
		&	\textcolor{black}{27.1}	&	\textcolor{black}{45.7}&	\textcolor{black}{53.6}&	\textcolor{black}{55.0}
		&	\textcolor{black}{26.0}&	\textcolor{black}{54.0}&	\textcolor{black}{68.0}&	\textcolor{black}{80.0} \\
		MS-SDALF\,\cite{farenzena2010person}	
		& 5.2 &	20.7 & 32.0	& 47.9
		& 6.3 &	18.8 & 27.1	& 37.3 
		& \textcolor{black}{36.4} & \textcolor{black}{47.1} & \textcolor{black}{59.3} & \textcolor{black}{67.9}
		& \textcolor{black}{46.0} & \textcolor{black}{52.0} & \textcolor{black}{74.0} & \textcolor{black}{86.0}
		\\
		\hline
		eSDC\cite{zhao2013unsupervised}
		&	25.8	&	43.6	&	52.6	&	62.0
		&	10.2	&	24.8	&	35.5	&	52.9 
		& - & - & - & -
		& - & - & - & - \\
		\hline
		MS-ColLBP
		&{34.3}	&{56.0}	&65.5 & 77.3
		&{23.2}	&{44.2}	&54.1 & {68.8} 
		& \textcolor{black}{47.9} & \textcolor{black}{56.4}& \textcolor{black}{60.0} & \textcolor{black}{66.4}
		& \textcolor{black}{34.0} & \textcolor{black}{50.0} & \textcolor{black}{68.0} & \textcolor{black}{84.0} \\
		\hline
		\textbf{DVR}
		& \textbf{40.0} & \textbf{71.7} & \textbf{84.5} & \textbf{92.2}
		& \textbf{39.5} & \textbf{61.1} &\textbf{71.7} & \textbf{81.0}
		&\textcolor{black}{\bf 54.3} & \textcolor{black}{\bf 70.0} & \textcolor{black}{\bf 77.9} & \textcolor{black}{\bf 85.0} 
		&\textcolor{black}{\bf 52.0} & \textcolor{black}{\bf 70.0} & \textcolor{black}{\bf 90.0} & \textcolor{black}{\bf 100} 
		\\	
		\hline
	\end{tabular}
	\vspace{-0.6cm}
\end{table*}

We compared the proposed DVR model with contemporary gait recognition
and temporal sequence matching methods for person (re-)identification.
(I) Gait recognition\,(GEI+RSVM)\,\cite{martin2012gait} is
a state-of-the-art gait recognition model using Gait Energy Image
(GEI)~\cite{Han2006GEI} 
(computed from pre-segmented silhouettes) 
as sequence representation and
RankSVM~\cite{chapelle2010efficient} for recognition.  
A challenge for applying gait recognition to unregulated image sequences in ReID scenarios is to generate good gait silhouettes as input. 
To that end, 
we first deployed the
DPAdaptiveMedianBGS algorithm provided by
the BGSLibrary~\cite{bgslibrary} to extract silhouettes from image
sequences given by each dataset.
This approach produces better 
foreground masking than other alternatives.
%
(II) ColLBP/HoGHoF/ColLBPHoGHoF+DTW applies
Dynamic Time Warping~\cite{rabiner1993fundamentals} 
to compute the similarity between two
sequences, using either ColLBP~\cite{hirzer2012relaxed}
or HoGHoF~\cite{laptev2008learning} or their combination as the per-frame 
feature descriptor.
This is similar to the approach of Simonnet \etal~\cite{simonnet2012},
except that they only used colour features. In
comparison, ColLBP is a stronger representation as it encodes
both colour and texture. Alternatively, HoGHoF encodes both texture
and motion information.

Table~\ref{tab:compare_alternatives}
presents the comparative ReID results among
DVR, GEI+RSVM (gait), ColLBP+DTW, HoGHoF+DTW,
and ColLBPHoGHoF+DTW. It is evident
that the proposed DVR outperforms significantly any competitor on all datasets.
	Gait recognition~\cite{martin2012gait}
	gives significantly weaker performance than
	the DVR model on every dataset.
	In comparison,
	its ReID accuracy on PRID$2011$ and HDA+ is much better 
	than that on iLIDS-VID.
This is because the GEI gait features are very sensitive to background
clutter and occlusions,
as shown by the examples in Fig.~\ref{fig:gei}.
It is obvious that the extracted gait
foreground masks from the iLIDS-VID person sequence
(middle) are contaminated more heavily by cluttered background and
other moving objects, compared to those from either PRID$2011$ (top)
or HDA+ (bottom).
Our DVR model trains itself 
by simultaneously selecting and ranking
only those video fragments which suffer
the least from occlusions and noise.
Moreover, DTW based sequence matching methods 
using either ColLBP, HoGHoF, or their combination also suffer notably from the inherently
uncertain nature of ReID sequences and perform significantly poorer than the proposed DVR 
approach. This is largely due to: 
(1) Person sequences have different
durations with arbitrary starting/ending frames, also potentially different 
numbers of
walking cycles.
Therefore, attempts to 
match entire sequences holistically
inevitably suffer from mismatching with erroneous similarity measurement;
(2) There is no clear (explicit) mechanism to avoid
incomplete/missing data, typical in crowded scenes;
(3) Direct sequence matching is less discriminative than 
learning an inter-camera discriminative
mapping function, which is explicitly built into the DVR model by
exploring multi-instance (fragment-pair) selection and ranking.

\subsection{Comparing Spatial Feature Representations}
\label{sec:comp_state-of-the-art}

To evaluate the effectiveness of discriminative video fragment selection
and ranking using both spatial appearance and space-time features 
for person ReID, we
compared the proposed DVR model against a wide range of contemporary
ReID models using spatial features, either in single-shot or 
multi-shot (multi-frames). In order to process the iLIDS-VID
dataset for our experiments, we mainly considered contemporary methods with
code available publicly. They
include (1) {\em SDALF}~\cite{farenzena2010person}~(single-/multi-shot versions);
(2) {\em eSDC}\footnote{
	The eSDC model\cite{zhao2013unsupervised}
	cannot be evaluated 
	on the small HDA+ dataset
	as it requires additionally
	saliency statistics modelling
	with two large reference sets which 
	are not available on HDA+.
}~\cite{zhao2013unsupervised};
(3) {\em SS-ColLBP} which uses RankSVM~\cite{chapelle2010efficient} as model
and colour\&LBP~\cite{hirzer2012relaxed} as representation;
(4) We also extended SS-ColLBP 
to multi-shot by averaging the ColLBP features of each frame over
an image sequence to focus on stable appearance cues and suppress
noise, in a similar approach to \cite{John2008}. We call this method
{\em MS-ColLBP}.
\textcolor{black}{
Moreover, we discuss the effect of clothing variation on person ReID methods,
a challenging topic which is mostly ignored and
under-investigated currently in the literature.
}

\vspace{.1cm}
\noindent \textcolor{black}{\em Comparing with spatial feature based methods} --
The results in Table~\ref{tab:compare_stateart} 
show that the proposed DVR model
outperforms significantly all the spatial feature based methods 
on all datasets,
\textcolor{black}{
e.g. 
 it gains $55.0\%$ and $287.3\%$ Rank-$1$ improvement over
eSDC; it also yields $16.6\%$, $70.3\%$, $13.4\%$ and $52.9\%$
Rank-$1$ improvement over MS-ColLBP 
on PRID$2011$, iLIDS-VID, HDA+($5$fps) and HDA+($2$fps) respectively}.
Note that the improvement margin achieved by the DVR model 
on iLIDS-VID (a more challenging dataset) is much more significant
than those on PRID$2011$ and \textcolor{black}{HDA+}. 
This demonstrates the effectiveness of the proposed
selective sequence matching method in coping with challenging
real-world data for learning a robust re-identification ranking function.
More concretely, the power of our DVR model can be largely attributed to 
identity-sensitive space-time gradient cues
learned by our discriminative fragment selection based matching and ranking mechanism,
beyond the conventional models of only learning from the spatial appearance data, e.g. colour and texture.

\begin{table*}[!t] \scriptsize
	\renewcommand{\arraystretch}{1.15}
	\caption{\footnotesize
		Complementary effect of the DVR model to existing spatial appearance
		feature based models. 
		MS: Multi-Shot.
	}
	\vskip -0.3cm
	\label{tab:complement_stateart}
	\centering
	\begin{tabular}
		{p{2.7cm}|
			p{.2cm}p{.2cm}p{.2cm}p{.3cm}|
			p{.2cm}p{.2cm}p{.2cm}p{.3cm}|
			p{.2cm}p{.2cm}p{.2cm}p{.3cm}|
			p{.2cm}p{.2cm}p{.2cm}p{.3cm}
		}
		\hline
		Dataset & 
		\multicolumn{4}{c|}{PRID$2011$\cite{hirzer11a}} & 
		\multicolumn{4}{c|}{iLIDS-VID\cite{TaiqingECCV14}} &
		\multicolumn{4}{c|}{\textcolor{black}{HDA+($5$fps)\cite{figueira2014hda}}} & 
		\multicolumn{4}{c}{\textcolor{black}{HDA+($2$fps)\cite{figueira2014hda}}} \\
		\hline
		Rank $R$ \textcolor{black}{(\%)}
		& 1 & 5 & 10 & 20 
		& 1 & 5 & 10 & 20 
		& \textcolor{black}{1} & \textcolor{black}{2} & \textcolor{black}{3} & \textcolor{black}{4}
		& \textcolor{black}{1} & \textcolor{black}{2} & \textcolor{black}{3}  & \textcolor{black}{4}
		\\
		\hline \hline 
		\textbf{DVR}
		& {40.0} & {71.7} & {84.5} & {92.2}
		& {39.5} & {61.1} & {71.7} & {81.0} 
		&\textcolor{black}{54.3} & \textcolor{black}{ 70.0} & \textcolor{black}{ 77.9} & \textcolor{black}{85.0} 
		&\textcolor{black}{ 52.0} & \textcolor{black}{ 70.0} & \textcolor{black}{ 90.0} & \textcolor{black}{\bf 100}
		\\	
		\hline
		MS-ColLBP
		& 34.3 &56.0 &{65.5} &77.3
		& {23.2} &{44.2} &{54.1} &{68.8} 
		& \textcolor{black}{47.9} & \textcolor{black}{56.4} & \textcolor{black}{60.0} & \textcolor{black}{66.4} 
		& \textcolor{black}{34.0} & \textcolor{black}{50.0} & \textcolor{black}{68.0} & \textcolor{black}{84.0} \\
		\textbf{MS-ColLBP+DVR}
		& 42.5 & 70.1 & 83.5 & 92.8 
		& 41.0 & 62.1 & \textbf{73.6} & 82.5
		&\textcolor{black}{\bf 59.3}	&	\textcolor{black}{74.3}&	\textcolor{black}{78.6}	&	\textcolor{black}{\bf 86.4}			
		&	\textcolor{black}{52.0}	&	\textcolor{black}{\bf 76.0}	&	\textcolor{black}{94.0}	&	\textcolor{black}{\bf 100} \\
		\hline
		MS-SDALF\cite{farenzena2010person}	
		&	5.2		&	20.7	&	32.0	&	47.9
		&	6.3		&	18.8	&	27.1	&	37.3 
		& \textcolor{black}{36.4} & \textcolor{black}{47.1} & \textcolor{black}{59.3} & \textcolor{black}{67.9}
		& \textcolor{black}{46.0} & \textcolor{black}{52.0} & \textcolor{black}{74.0} & \textcolor{black}{86.0}\\
		\textbf{MS-SDALF+DVR}
		&	{44.2}	&	{71.2}	&	{85.1}	&	{92.5}
		&	{40.9}	&	{62.7}	&	{72.1}	&	{82.1}
		&	\textcolor{black}{\bf 59.3}&\textcolor{black}{\bf 75.7}&	\textcolor{black}{\bf 82.9}&\textcolor{black}{ 85.0}		
		&	\textcolor{black}{\bf 54.0}	&	\textcolor{black}{\bf 76.0}	&	\textcolor{black}{\bf 98.0}	&	\textcolor{black}{\bf 100} \\
		\hline
		eSDC\cite{zhao2013unsupervised}
		&	25.8	&	43.6	&	52.6	&	62.0
		&	10.2	&	24.8	&	35.5	&	52.9  
		& - & - & - & -
		& - & - & - & -
		\\
		\textbf{eSDC+DVR}
		& 48.2 & \textbf{75.2} & {87.0}& 94.2
		& 41.0 & \textbf{63.7} & 72.7 & \textbf{83.3} 
		& - & - & - & -
		& - & - & - & -
		\\
		\hline
		eSDC+MS-SDALF
		& 25.1 & 42.9 & 52.0 & 62.2
		& 10.2 & 25.3 & 35.2 & 52.9 
		& - & - & - & -
		& - & - & - & - \\
		\textbf{eSDC+MS-SDALF+DVR}
		& \textbf{48.3} & 74.9 & \textbf{87.3} & \textbf{94.4}
		& \textbf{41.3} & 63.5 & 72.7 & 83.1 
		& - & - & - & -
		& - & - & - & - \\	
		\hline
	\end{tabular}
	\vspace{-0.2cm}
\end{table*}


\begin{table*}[!t] \scriptsize
	\renewcommand{\arraystretch}{1.15}
	\caption{\footnotesize
		The effect of space-time video fragment selection. SS: Single-Shot; MS: Multi-Shot.
	}
	\vskip -0.3cm
	\label{tab:selection_effect}
	\centering
	\begin{tabular}
		{l|p{.2cm}p{.2cm}p{.2cm}p{.3cm}|
			p{.2cm}p{.2cm}p{.2cm}p{.3cm}|
			p{.2cm}p{.2cm}p{.2cm}p{.3cm}|
			p{.2cm}p{.2cm}p{.2cm}p{.3cm}
		}
		\hline
		Dataset & 
		\multicolumn{4}{c|}{PRID$2011$\cite{hirzer11a}} & \multicolumn{4}{c|}{iLIDS-VID\cite{TaiqingECCV14}}  &
		\multicolumn{4}{c|}{\textcolor{black}{HDA+($5$fps)\cite{figueira2014hda}}} & 
		\multicolumn{4}{c}{\textcolor{black}{HDA+($2$fps)\cite{figueira2014hda}}}\\
		\hline
		Rank $R$ \textcolor{black}{(\%)}
		& 1 & 5 & 10 & 20
		& 1 & 5 & 10 & 20
		& \textcolor{black}{1} & \textcolor{black}{2} & \textcolor{black}{3} & \textcolor{black}{4} 
		& \textcolor{black}{1} & \textcolor{black}{2} & \textcolor{black}{3}  & \textcolor{black}{4} \\
		\hline
		\hline
		SS-ColHOG3D		
		&	25.7	&	49.1	&	61.0	&	75.5
		&	15.5    &	33.7	&	47.5	&	61.4
		&	\textcolor{black}{40.7}	&	\textcolor{black}{57.1}&	\textcolor{black}{64.3}&	\textcolor{black}{72.9}
		&	\textcolor{black}{30.0}	&	\textcolor{black}{54.0}&	\textcolor{black}{78.0}&	\textcolor{black}{90.0} \\
		MS-ColHOG3D	
		&	29.6	&	54.9	&	70.8	&	86.1
		&	19.9	&	42.4	&	53.6	&	67.6
		&	\textcolor{black}{50.7}&	\textcolor{black}{65.0}&	\textcolor{black}{68.6}	&	\textcolor{black}{73.6}
		&	\textcolor{black}{30.0}&	\textcolor{black}{60.0}&	\textcolor{black}{\bf 90.0}&	\textcolor{black}{96.0} \\
		
		\hline
		\textbf{DVR}	
		& \textbf{40.0} & \textbf{71.7} & \textbf{84.5} & \textbf{92.2}
		& \textbf{39.5} & \textbf{61.1} &\textbf{71.7} & \textbf{81.0} 
		&\textcolor{black}{\bf 54.3} & \textcolor{black}{\bf 70.0} & \textcolor{black}{\bf 77.9} & \textcolor{black}{\bf 85.0} 
		&\textcolor{black}{\bf 52.0} & \textcolor{black}{\bf 70.0} & \textcolor{black}{\bf 90.0} & \textcolor{black}{\bf 100}
		\\	
		\hline
	\end{tabular}
	\vspace{-0.5cm}
\end{table*}

\begin{figure}[!t]
	\centering
	\includegraphics[width=0.9\linewidth]{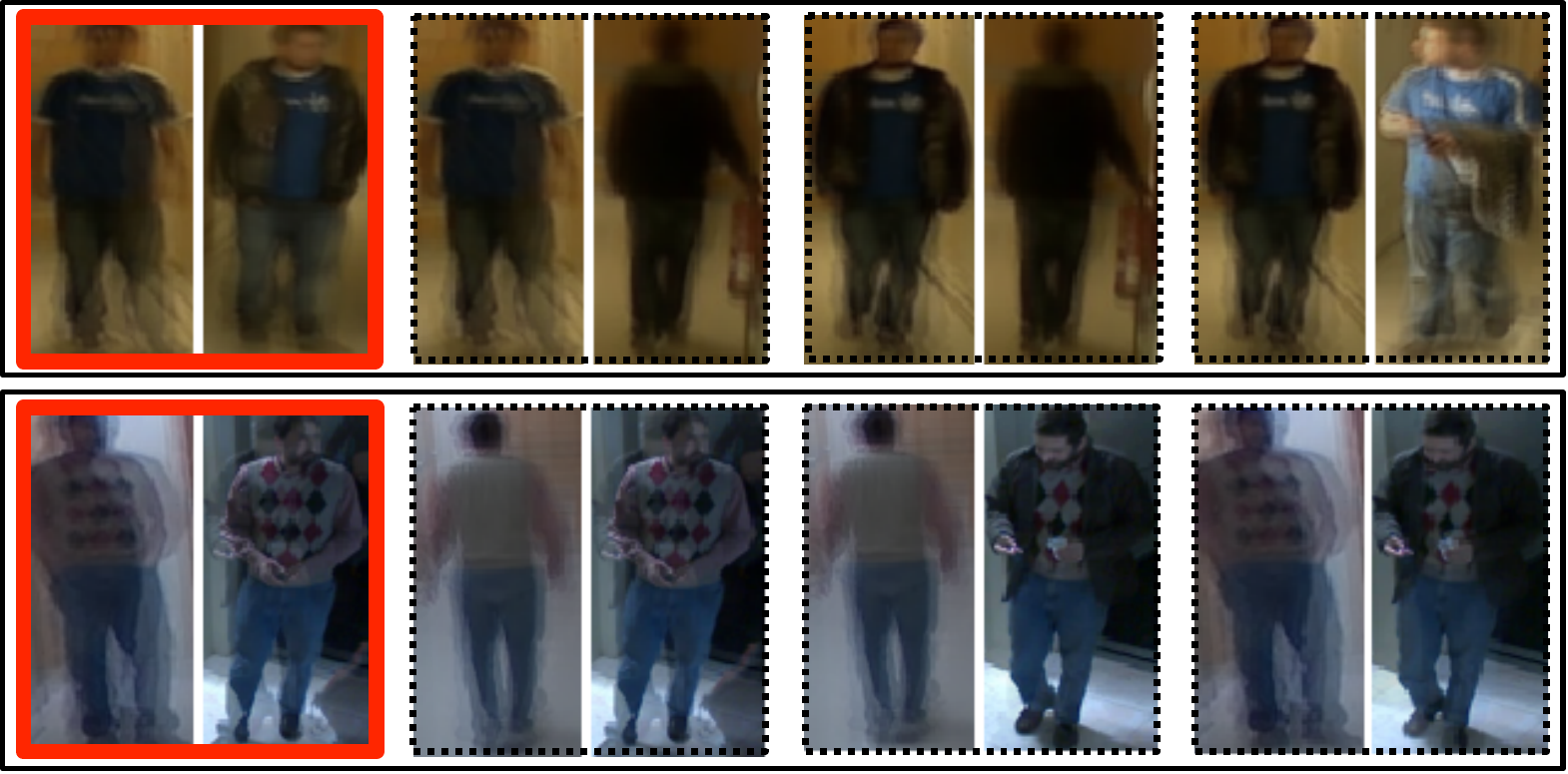}%
	\vskip -.3cm
	\caption{
	    \footnotesize
		Examples of automatically selected cross-camera fragment pairs (indicated with red bounding box) from the HDA+ dataset during person ReID by our DVR model.
		Top row: from HDA+($5$fps); Bottom row: from HDA+($2$fps).
		A fragment is visualised as the weighted average of all its frames with emphasis on its central frame.
	}
	\label{fig:cloth_change}
	\vspace{-.5cm}
\end{figure}

\vspace{.1cm}
\noindent
{\em Clothing change challenge} -- 
Existing person ReID studies typically assume 
no changes
in
clothing.
However, this assumption is not always valid. Realistically,
clothing may change for some people within and/or across camera
views.
Specifically,
while there is no ($0\%$) explicit clothing change among the
people in both PRID$2011$ and iLIDS-VID,
$35.7\%$ people changed their jacket/coat/shirt
in HDA+($5$fps) and $50.0\%$ in HDA+($2$fps), resulting in 
substantial change in appearance
(Fig.~\ref{fig:dataset_example}(c,d)). Whilst only partial
appearance variation may arise from changes in
viewpoint and lighting, severe occlusion
can also cause significant appearance change
(Fig.~\ref{fig:dataset_example}(a,b)).
Given this observation, we compared the performance of DVR
against other appearance-based ReID models on the
four different datasets with different degrees of
clothing changes. 
We pay special
attention to multi-shot models as they are expected to
be more robust
under clothing changes. 
The results in Table \ref{tab:compare_stateart} show that
MS-SDALF benefits consistently from multiple shots on all four
datasets, either with clothing changes or not.  
This is 
largely due to
its body-part selective matching strategy, 
i.e. using the
best-matched
patch pairs during matching.
However, this method can also give weak ReID accuracy due to the inherent difficulties in obtaining explicitly reliable body-part segmentation in surveillance images. 
In comparison, MS-ColLBP suffers considerably more from
clothing changes, evident from a decreased performance
advantage over SS-ColLBP on HDA+($5$fps) and worse still on
HDA+($2$fps), when compared with those on PRID$2011$ and iLIDS-VID.
This suggests that the advantage of MS-ColLBP over SS-ColLBP decreases
when clothing changes are abrupt at low frame rates.
Under such conditions, averaging without selection is a poor
strategy to cope with clothing changes. 
In contrast, the proposed DVR model 
not only explores discriminative space-time ReID information less
sensitive to appearance change, but also selects automatically
the best-matched fragments for appearance
consistency, sharing a similar principle of MS-SDALF but
being
more flexible and robust without requiring explicit 
part
segmentation. 
	We show in Fig. \ref{fig:cloth_change} two examples of model
        selected discriminative fragment pairs across camera views for
        person ReID. 
	Note, this selection is driven by both static appearance and
        dynamic motion information embedded in our DVR model design.
This demonstrates the potential advantage of the DVR
model in addressing the clothing change challenge in person
ReID, a problem under-studied in the current literature. 

\subsection{Complementary to Spatial Features}
\label{sec:complement_state-of-the-art}

We further evaluated the complementary effect between the DVR model
and existing colour/texture feature based ReID approaches. 
The results are reported in 
Table~\ref{tab:complement_stateart}. 
It is evident that for any existing appearance model, significant
performance gain is achieved by incorporating the DVR ranking score
(Eqn.~(\ref{eqn:combination})) into its ranking result. 
More specifically,
\textcolor{black}{ 
on PRID$2011$ and iLIDS-VID, the Rank-$1$ ReID performance
of using multi-shot colour and texture features (MS-ColLBP) is boosted by
$23.9\%$ and $76.7\%$;}
Rank-$1$ of eSDC is improved by $86.8\%$ and $302.0\%$;
Rank-$1$ of eSDC+MS-SDALF 
is increased by $92.4\%$ and $304.9\%$, respectively.
Similar improvements are gained on low frame rate sequences from HDA+ by MS-ColLBP and MS-SDALF.
Such a performance step-change in improving
conventional spatial feature based models is primarily
due to the
exploration of discriminative space-time features and the fragment selection based matching scheme 
by the proposed DVR model. 
This space-time selective matching process discovers
mostly independent source of information when comparing with all static
appearance features, therefore playing a significant complementary
and beneficial role to contemporary spatial feature based models.
It is also worth pointing out that most existing spatial feature based
methods benefit more from combining with DVR when tested on iLIDS-VID, 
and less on PRID$2011$ and HDA+.
This observation highlights the importance and necessity of 
discriminative fragment selection for robust model learning  
given video data from more crowded public scenarios where
blind learning from all the sequence data without selection leads to
poorer and degraded models.

In addition, it is evident from Table~\ref{tab:complement_stateart} 
that the DVR model can benefit from 
combining with other spatial feature based ReID models, although slightly.
This gain may be explained as the result of drawing 
from
diverse sources of spatial features.

\subsection{Evaluation of Space-time Fragment Selection}
\label{sec:exp_selection}

To evaluate the space-time video fragment selection mechanism in the proposed DVR
model, we implemented two baseline methods without this selection mechanism: 
(1) SS-ColHOG3D represents
each image sequence by ColHOG3D 
features of a single fragment randomly selected from the image sequence; 
(2) MS-ColHOG3D represents each image sequence by the averaged
ColHOG3D features of four fragments uniformly selected from the
sequence.
In both baseline methods, 
RankSVM~\cite{chapelle2010efficient} is used to rank the person
sequence representations.
For a fair comparison, the length of these fragments used for both baselines is set the same as 
that in our DVR model.

The results are presented in Table~\ref{tab:selection_effect}.
The DVR model outperforms
SS-ColHOG3D and MS-ColHOG3D in Rank-$1$ 
by $55.6\%$ and $35.1\%$ on PRID$2011$,
by $33.4\%$ and $7.1\%$ on HDA+($5$fps), 
and by $73.3\%$ and $73.3\%$ on HDA+($2$fps).
The performance advantage of DVR over SS-ColHOG3D and MS-ColHOG3D is
even greater on the more challenging iLIDS-VID dataset, i.e. yielding $154.8\%$
and $98.5\%$ Rank-$1$ improvement respectively.
This demonstrates clearly that in the presence of significant noise and
given unregulated person image sequences,
it is indispensable to automatically select discriminative
space-time fragments from raw image sequences in order to construct a
more robust model for person ReID.
It is also noted that MS-ColHOG3D
outperforms SS-ColHOG3D by
suppressing noise using temporal averaging.
Although such a straightforward averaging approach can have some
benefits over single-shot methods, it loses out on discriminative
information selection due to uniform temporal smoothing.

\section{Conclusion and Future Work}

\textbf{Conclusion} 
We have presented a novel DVR framework for person re-identification
by video ranking using discriminative space-time and appearance
feature selection. Our extensive evaluations show
that this model outperforms a wide range of contemporary techniques
from gait recognition and temporal sequence matching to state-of-the-art
single-/multi-shot(or frame) spatial feature representation
based ReID models.
In contrast to existing ReID approaches 
that often employ spatial appearance of people alone, 
the proposed method is capable of 
capturing more accurately both appearance and space-time information 
discriminative for person ReID 
through learning a cross-view multi-instance ranking function.
This is made possible by the ability of our model to
discover and exploit automatically the most reliable and informative video fragments
extracted from inherently incomplete and inaccurate person image sequences
captured against cluttered backgrounds, without any guarantee on
person walking cycles, starting/ending frame alignment,
video frame rates, and clothing stability.
Moreover, the proposed DVR model significantly complements and improves
existing spatial appearance features when combined for
person ReID. 
Extensive comparative evaluations were conducted to validate the
advantages of the proposed model over a variety of baseline methods
on three challenging image sequence based ReID datasets.

\vspace{0.1cm}
\noindent \textbf{Future work} 
Person re-identification remains largely an unsolved problem~\cite{REIDChallenge:14}, and our future work includes: 
(1) In addition to space-time information, how to exploit automatically other knowledge sources,
e.g. the topology structure of a camera network, 
or the semantic description (e.g. mid-level attributes nameable by human) 
of people's appearance and walking style;
(2) How to cope with open-world person re-identification settings~\cite{Shaogang2014reid,zhengtowards,braisCRF}
where the probe people are not guaranteed to appear in the gallery set.

\vspace{-.1cm}
\section*{Acknowledgement}
\footnotesize
	We shall thank Dario Figueira of IST for providing the
        HDA+ dataset and for assisting in extracting the person
        bounding boxes from raw videos required for person ReID evaluations
        and gait experiments;
        Martin Hirzer, Peter Roth and Csaba Beleznai of AIT for
        providing the additional raw videos of PRID$2011$
        required for gait recognition experiments.
        Corresponding authors: Shaogang Gong and Shengjin Wang.
\vspace{-.1cm}

\ifCLASSOPTIONcaptionsoff
  \newpage
\fi

\bibliographystyle{IEEEtran}
\bibliography{TPAMI14_ReId}

%

\vspace{-0.5cm}
\begin{biography} [{\includegraphics[width=1in,height=1.25in,clip,keepaspectratio]{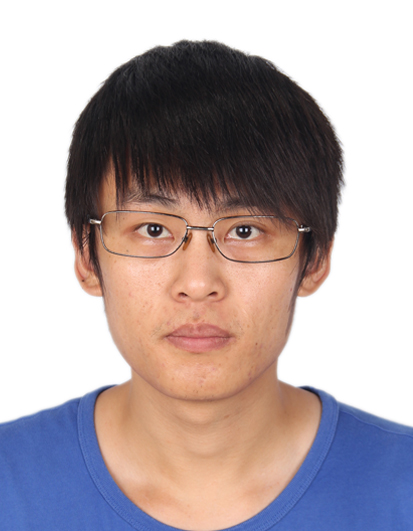}}]{Taiqing Wang}
received his B.E. (2009) from Beijing University of Posts and
Telecommunications and his Ph.D. (2015) from Tsinghua
University. His research interests include computer
vision, pattern recognition and machine learning. 
\end{biography}
\vspace{-1.2cm}

\begin{biography}[{\includegraphics[width=1in,height=1.25in,clip,keepaspectratio]{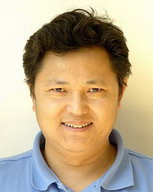}}]{Shaogang Gong}
is Professor of Visual Computation
at Queen Mary University of London (since 2001), a
Fellow of the Institution of Electrical Engineers
and a Fellow of the British Computer Society.
He received his D.Phil (1989) in computer vision from
Keble College, Oxford University. His
research interests include computer vision, machine
learning and video analysis.
\end{biography}
\vspace{-1.2cm}

\begin{biography}[{\includegraphics[width=1in,height=1.25in,clip,keepaspectratio]{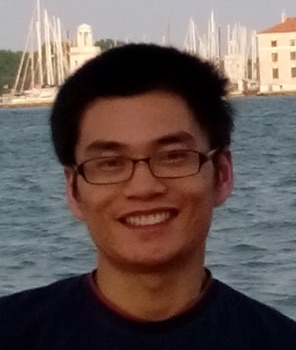}}]{Xiatian Zhu}
received his B.Eng. and M.Eng. 
from University of Electronic Science and Technology of China, and
his Ph.D. (2015) from Queen Mary University of London. 
His research interests include computer vision, 
pattern recognition and machine learning.
\end{biography} 
\vspace{-1.2cm}

\begin{biography}[{\includegraphics[width=1in,height=1.25in,clip,keepaspectratio]{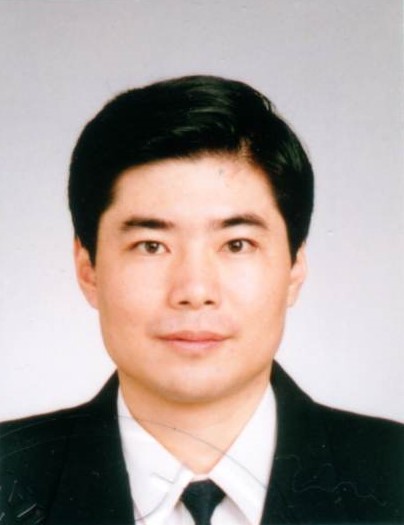}}]
{Shengjin Wang} is Professor of Electronic Engineering at Tsinghua
University (since 2003). He received his B.E. (1985) from Tsinghua
University and his Ph.D. (1997) from Tokyo Institute of
Technology. From 1997 to 2003, he was a Member of the
Senior Research Staff in the Internet System Research Laboratories,
NEC Corporation, Nara, Japan. His current
research interests include computer vision, video surveillance, and
virtual reality. 
\end{biography}





\end{document}